\documentclass[conference]{IEEEtran}
\usepackage{cite}
\usepackage{amsmath,amssymb,amsfonts}
\usepackage{graphicx}
\usepackage{textcomp}
\usepackage{xcolor}
\def\BibTeX{{\rm B\kern-.05em{\sc i\kern-.025em b}\kern-.08em
    T\kern-.1667em\lower.7ex\hbox{E}\kern-.125emX}}

\usepackage{enumitem}
\usepackage{hyperref} 
\usepackage{graphicx}
\usepackage{booktabs}
\usepackage{amsmath}
\usepackage{amsthm}
\usepackage{bbm}
\usepackage{amsfonts}
\usepackage{epsfig}
\usepackage{url}
\usepackage{diagbox}
\usepackage{balance}
\usepackage{algorithm}
\usepackage{algorithmicx}
\usepackage[noend]{algpseudocode}
\usepackage{makecell}
\usepackage{multirow}
\usepackage{xspace}
\usepackage{graphicx} 
\usepackage{caption}
\usepackage{tabularray}
\usepackage{subcaption}
\usepackage{cite}
\usepackage{mathtools}
\usepackage{tabularx}
\usepackage{empheq}
 
\usepackage{textcomp}
\usepackage{xcolor}

\usepackage{subcaption}
\def\BibTeX{{\rm B\kern-.05em{\sc i\kern-.025em b}\kern-.08em
T\kern-.1667em\lower.7ex\hbox{E}\kern-.125emX}}

\setlist{topsep=0pt,noitemsep} \setitemize[1]{label=$\circ$}

\newcommand{\eat}[1]{}
\newcommand{\stab}{\rule{0pt}{8pt}\\[-1.6ex]}

\newcommand{\sstab}{\rule{0pt}{8pt}\\[-2.4ex]}

\newcommand{\bi}{\begin{itemize}}
\newcommand{\ei}{\end{itemize}}
\newcommand{\tbi}{
  \begin{itemize}[leftmargin=10pt, itemsep=0pt, topsep=0pt, parsep=0pt, partopsep=0pt]
}

\newcommand{\be}{\begin{enumerate}}
\newcommand{\ee}{\end{enumerate}}
\newcommand{\beqn}{\begin{eqnarray*}}
\newcommand{\eeqn}{\end{eqnarray*}}

\newcommand{\stitle}[1]{
 \vspace{1.5ex}
\noindent{\bf #1}}

\newcommand{\ie}{\emph{i.e.,}\xspace}
\newcommand{\eg}{\emph{e.g.,}\xspace}
\newcommand{\wrt}{\emph{w.r.t.}\xspace}

\newcommand{\kwlog}{\emph{w.l.o.g.}\xspace}

\DeclareMathOperator*{\argmin}{arg\,min}

\newcommand{\kw}[1]{{\ensuremath {\mathsf{#1}}}\xspace}

\newcounter{ccc}

\newcommand{\NP}{\kw{NP}}

\newcommand{\PTIME}{\kw{PTIME}}

\newcommand{\M}{{\mathcal M}}
\newcommand{\R}{{\mathcal R}}

\newcommand{\G}{{\mathcal G}}

\newcommand{\eop}{\hspace*{\fill}\mbox{$\Box$}}     
\newcounter{example}
\renewcommand{\theexample}{\arabic{example}}
\newenvironment{example}{
         \vspace{1.5ex}
        \refstepcounter{example}
        {\noindent\bf Example \theexample:}}{
        \eop
         \vspace{1.5ex}
        }

\newcommand{\nthesection}{\arabic{section}}
\newcounter{theorem}
\renewcommand{\thetheorem}{\arabic{theorem}}
\newcounter{prop}
\renewcommand{\theprop}{\arabic{theorem}}
\newcounter{lemma}
\renewcommand{\thelemma}{\arabic{theorem}}
\newcounter{cor}
\renewcommand{\thecor}{\arabic{theorem}}
\newenvironment{theorem}{\begin{em}
        \refstepcounter{theorem}
        {
         \vspace{1.5ex}
        \noindent\bf  Theorem  \thetheorem:}}{
        \end{em}\eop
         \vspace{1.5ex}
        } 
\newenvironment{lemma}{\begin{em}
        \refstepcounter{theorem}
        {
         \vspace{1ex}
        \noindent\bf Lemma \thelemma:}}{
        \end{em}\eop
         \vspace{1ex}
        } 
\newcounter{alg}[section]
\renewcommand{\thealg}{\nthesection.\arabic{alg}}

\newcounter{arule}
\renewcommand{\thearule}{\arabic{arule}}

\newcounter{claim}

\newenvironment{proofS}{
         \vspace{1ex}
        {\noindent\bf Proof sketch:\ }}{\eop
         \vspace{1ex}
        }
\renewcommand{\texttt}[1]{{\small\textsf{#1}}}

\newcommand{\warn}[1]{{\color{red}{#1}}}

\DeclareMathOperator*{\argmax}{arg\,max}

\newcommand{\gnn}{\kw{GNN}}

\newcommand{\eetitle}[1]{
 \vspace{0.8ex}
\noindent{\em\underline{#1}}}

\newcommand{\gnns}{\kw{GNNs}}
\newcommand{\gat}{\kw{GAT}}

\newcommand{\gcn}{\kw{GCN}}
\newcommand{\gcns}{\kw{GCNs}}

\newcommand{\gin}{\kw{GIN}}

\newcommand{\arxiv}{\kw{OGBN\_arxiv}}

\newcommand{\cora}{{\kw{Cora}}}
\newcommand{\pubmed}{{\kw{PubMed}}}

\newcommand{\facebook}{{\kw{FacebookPage}}} 
\newcommand{\amazoncomputer}{{\kw{AmazonComputer}}}
\newcommand{\gnnexp}{{\kw{GExp}}}

\newcommand{\pgexp}{{\kw{PGExp}}}
\newcommand{\cff}{\kw{CF\textsuperscript{2}}}

\newcommand{\moe}{\kw{MOExp}}

\newcommand{\bahouse}{\kw{BA\_Shapes}}
\newcommand{\treecycle}{\kw{Tree\_Cycles}}

\newcommand{\gset}{{\mathcal G}}
\newcommand{\drg}{{\em DRG}}

\newcommand{\dscore}{{\sf DS}}
\newcommand{\dset}{{\mathcal D}}
\newcommand{\eset}{{\mathcal E}}
\newcommand{\divs}{{\sf DivS}}
\newcommand{\ncs}{{\sf NCS}}
\newcommand{\cd}{{\sf CD}}

\newcommand{\veriF}{{\sf vrfyF}}
\newcommand{\veriCF}{{\sf vrfyCF}}

\newcommand{\sover}{\overline{s}}
\newcommand{\as}{{\sf AS}}

\newcommand{\updatesx}{{\sf updateSX}\xspace}
\newcommand{\updatedivsx}{{\sf updateDSX}\xspace}
\newcommand{\swap}{{\sf swap}\xspace}

\newcommand{\diveval}{\kw{DivEVAL}}

\newcommand{\dq}[1]{{\color{blue}{#1}}}

\newcommand{\parasx}{\kw{ParaSX}}
\newcommand{\parasxn}{\kw{ParaSX}-\kw{N}}
\newcommand{\parasxeis}{\kw{ParaSX}-\kw{EIS}}
\newcommand{\cp}{\kw{CP}}

\newcommand{\gq}{{\mathcal G_Q}}

\newcommand{\rthree}[1]{{\color{blue}{#1}}}
\newcommand{\rfour}[1]{{\color{purple}{#1}}}

\newcommand{\acc}{\kw{acc}}

\newcommand{\sgxq}{\kw{SXQ}}
\newcommand{\sgxqs}{\kw{SXQs}}
\newcommand{\fac}{\kw{fac}}
\newcommand{\cfac}{\kw{cfac}}
\newcommand{\fplus}{\kw{fdl}^+}
\newcommand{\fminus}{\kw{fdl}^-}
\newcommand{\conc}{\kw{conc}}
\newcommand{\shap}{\kw{shapley}}

\newcommand{\eval}{\kw{EVAL}}
\newcommand{\verify}{\kw{Verify}}

\newcommand{\apxsx}{\kw{ASX}-\kw{OP}}
\newcommand{\apxsxi}{\kw{ASX}-\kw{I}}
\newcommand{\divsx}{\kw{DSX}}

\newcommand{\yw}[1]{{\color{red}{#1}}}

\begin{document}
\bstctlcite{IEEEexample:BSTcontrol}

\title{Interpreting Graph Inference with Skyline Explanations}

\author{
    \IEEEauthorblockN{Dazhuo Qiu}
    \IEEEauthorblockA{
        \textit{Aalborg University}\\
        Aalborg, Denmark \\
        dazhuoq@cs.aau.dk
    }
\and
    \IEEEauthorblockN{Haolai Che}
    \IEEEauthorblockA{
        \textit{Case Western Reserve University}\\
        Cleveland, USA \\
        hxc859@case.edu
    }
\and
    \IEEEauthorblockN{Arijit Khan}
        \IEEEauthorblockA{
        \textit{BGSU}, Bowling Green, USA\\
        \textit{AAU}, Aalborg, Denmark \\
        arijitk@bgsu.edu
    }
\and
    \IEEEauthorblockN{Yinghui Wu}
    \IEEEauthorblockA{
        \textit{Case Western Reserve University}\\
        Cleveland, USA \\
        yxw1650@case.edu
    }
}

\maketitle

\begin{abstract}
Inference queries have been routinely issued to graph machine learning models such as graph neural networks (\gnns) for various network analytical tasks. 
Nevertheless, \gnn outputs are often hard to interpret comprehensively. Existing methods typically conform to individual pre-defined explainability measures (such as fidelity), which often 
leads to biased, ``one-sided'' interpretations. 
This paper introduces {\em skyline explanation}, a new 
paradigm that interprets \gnn outputs by simultaneously optimizing multiple explainability measures of users' interests.  
(1) We propose skyline explanations as a 
Pareto set of explanatory subgraphs that 
dominate others over multiple 
explanatory measures. 
We formulate skyline explanation as a multi-criteria optimization problem, and establish its hardness results. 
(2) We design efficient algorithms with an onion-peeling approach, 
which strategically prioritizes nodes and removes unpromising edges 
to incrementally assemble 
skyline explanations. 
(3) We also develop an algorithm to diversify the skyline 
explanations to enrich the comprehensive interpretation. 
(4) We introduce efficient parallel algorithms with 
load-balancing strategies to scale skyline 
explanation for large-scale \gnn-based inference. 
Using real-world and synthetic graphs, we experimentally verify our algorithms' effectiveness and scalability. 
\end{abstract}



\section{Introduction}
\label{sec:intro}

Graph models such as graph neural networks (\gnns) 
have been trained and routinely queried to 
perform network analysis in, \eg biochemistry, social and financial networks \cite{Wang2023, you2018graph, cho2011friendship,wei2023neural}, 
among other graph analytical tasks. 
Given a graph $G$ and a set of test nodes $V_T$ in $G$, a \gnn $\M$ 
can be considered as a function 
that converts a feature representation 
of $G$ in the form of $(X,A)$, where 
$X$ (resp. $A$) refers to the node feature matrix (resp. 
normalized adjacency matrix) of $G$, 
to an output representation $Z$ (``output embedding matrix''). 
The output $Z$ can be post-processed to 
task-specific output such as 
labels for classification. 
An ``inference query'' invokes 
the above process to query $\M$, to perform the 
inference analysis of $G$. 
\eat{
Given a graph $G$ (a network representation of 
a real-world dataset), a \gnn $\M$
aims to 
learn the node representations of $G$ 
that can be converted to proper 
results for the targeted downstream 
tasks, e.g., node classification, link prediction, or regression analysis.  
For example, 
a \gnn-based node classification  
assigns a class label to a set of test nodes in $G$,  
where the label of each 
test node $v$ (the ``output'' of $\M$ at node $v$, denoted as $\M(v, G)$) is 
determined by the node representation learned by the \gnn $\M$. 
\gnns have been applied for node classification in biochemistry, social and financial networks \cite{Wang2023, you2018graph, cho2011friendship,wei2023neural}, 
among other graph analytical tasks. 
}

Despite the promising performance of \gnns, 
it remains desirable yet nontrivial to interpret their outputs 
to help users understand \gnn-based decision making~\cite{yuan2022explainability}. 
Several methods (``explainers'') have been proposed  ~\cite{ying2019gnnexplainer, lucic2022cf, tan2022learning, zhang2024gear, luo2020parameterized}. 
Typically, a \gnn explainer
extracts a subgraph $G_\zeta$ of $G$ 
that can best clarify the output of an inference query 
posed on $\M$ over $G$.  
This is often addressed 
by discovering a subgraph 
$G_\zeta$ subject to a pre-defined  
metric, which quantifies the explainability 
of $G_\zeta$ for the output (as summarized in Table~\ref{tab-measures}). 


For example, a subgraph $G_\zeta$ of a graph 
$G$ is a ``\textit{factual}'' 
explanation for $\M$ over $G$,  
if it preserves the output of $\M$ 
(hence is ``faithful'' to 
the output of $\M$)~\cite{ying2019gnnexplainer, luo2020parameterized,liu2021multi,tan2022learning}.  
$G_\zeta$  is a ``\textit{counterfactual}'' 
explanation, if removing 
the edges of $G_\zeta$ from $G$ leads to a change of 
the output of $\M$ on the remaining 
graph (denoted as $G\setminus G_\zeta$)~\cite{lucic2022cf, liu2021multi,tan2022learning}. Other 
metrics include $\kw{fidelity}^-$~\cite{lucic2022cf, liu2021multi,tan2022learning}  
(resp. $\kw{fidelity}^+$~\cite{luo2020parameterized, yuan2021explainability}), which quantifies 
the explainability of $G_\zeta$ in terms of 
the closeness between the task-specific  
output of $\M$, such as the probability 
of label assignments, over $G_\zeta$ (resp. $G\setminus G_\zeta$) and their original 
counterpart $G$, and ``\textit{conciseness} (\textit{sparsity})''
that favors small explanatory subgraphs. 

Nevertheless, such a 
metric assesses explanations from 
``one-sided'' perspective of explainability (\S~\ref{sec-pre}).
For example, an explanation that achieves high fidelity 
may typically ``compromise'' in conciseness, due to the need 
of including more nodes and edges from the original 
graphs to be ``faithful'' to the \gnn output.  
Consider the following example.

\begin{figure}[tb!]
    \centering
    \includegraphics[width=0.46\textwidth]{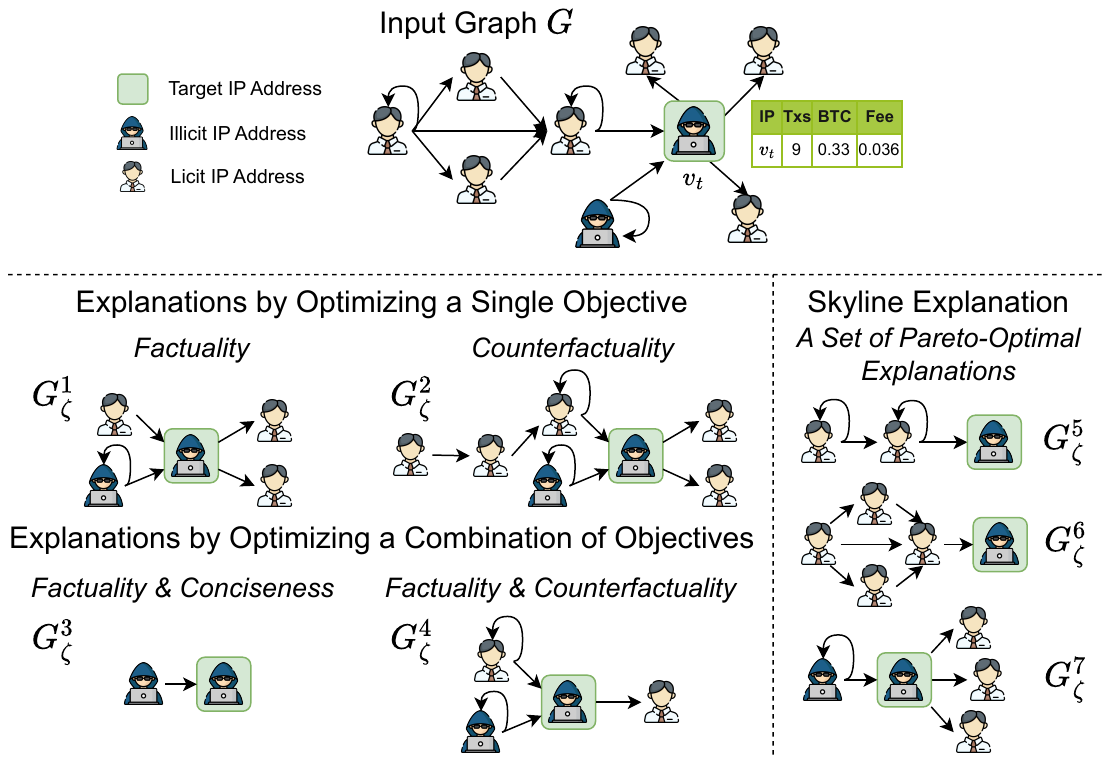}
       \vspace{-1ex}
    \caption{
    {\small 
    A Bitcoin transaction network with a target IP address (test node $v_t$) that has a label ``{\em illicit}'' to be explained~\cite{ellipticpp}. 
    }
    }
    \label{fig:motivation}
    \vspace{-4ex}
\end{figure}

\begin{table*}[tb!]
    \centering
    \caption{Representative explainability measures and notable \gnn explainers
    }
    \resizebox{2.0\columnwidth}{!}{
    \begin{tabular}{c|c|c|c|c|c} 
      {\bf Symbol}   & {\bf Measure} &  {\bf Equation}  & {\bf Range} & {\bf Description} & {\bf Explainers} \\
      \hline
      \fac   & factual  & $\M(v,G)$ = $\M(v,G_\zeta)?$ & \{\kw{true},\kw{false}\} & a Boolean function & ~\cite{ying2019gnnexplainer,luo2020parameterized,liu2021multi,tan2022learning} \\ 
      \cfac   & counterfactual & $\M(v,G)\neq \M(v,G\setminus G_\zeta)?$ & \{\kw{true},\kw{false}\} & a Boolean function & ~\cite{lucic2022cf, liu2021multi,tan2022learning}  \\
      $\fplus$
      &  fidelity$^+$ & $Pr(\M(v, G))-Pr(\M(v, G\setminus G_\zeta))$ & $[-1,1]$ & the larger, the better & ~\cite{chen2024view, yuan2021explainability} \\
      
      $\fminus$
      & fidelity$^-$ & $Pr(\M(v, G))-Pr(\M(v, G_\zeta))$ &  $[-1,1]$ & the smaller, the better & ~\cite{chen2024view, luo2020parameterized} \\
      
      \conc
      & conciseness & 
     $\frac{1}{N}\sum_{i=1}^{N}(1- 
     \frac{|E(G_\zeta)|}{|E(G)|})$

      & $[0,1]$ &  the smaller, the better & ~\cite{luo2020parameterized, yuan2021explainability} \\
      \shap 
      & Shapley value & $\phi
      (G_\zeta) = \sum_{S \subseteq P \setminus \{G_\zeta \}} \frac{|S|!(|P| - |S| - 1)!}{|P|!}m(S,G_\zeta)$ & $[-1,1]$ & total contribution of nodes in $G_\zeta$  & ~\cite{yuan2021explainability} \\ 
     \hline
    \end{tabular}
    }
    \vspace{-2ex}
    \label{tab-measures}
\end{table*}

\begin{example}
\label{exa-motivation}
Figure~\ref{fig:motivation} illustrates a fraction of 
a Bitcoin blockchain transaction network $G$. 
Each IP address (node) is associated with transactions (edges) 
and transaction features, such as frequency of transactions, the amount of transacted Bitcoins, and the amount of fees in Bitcoins. 
A \gnn-based classifier $\M$ is trained to detect illicit IP addresses, 
by assigning a label (\eg ``illicit'') to 
nodes (\eg $v_t$). 

\vspace{.5ex}
A law enforcement agency has posed an inference query 
over $\M$ to get the output, and 
wants to further understand ``why'' the \gnn 
asserts $v_t$ as an illicit account address. They may further ask 
clarifications by asking ``{\em which fraction of the graph $G$ is responsible for the \gnn's decision on assigning the label ``illicit'' to the account $v_t$?}'', and ground this output by referring to known real-world money laundering scenarios~\cite{spindle,peel}). 
For example, ``\textit{Spindle}''~\cite{spindle} suggests that perpetrators generate multiple shadow addresses to transfer small amounts of assets along lengthy paths to a specific destination;  
and \textit{Peel Chain}~\cite{peel} launders large amounts of cryptocurrency through sequences of small transactions, where minor portions are `peeled' from the original address and sent 
for conversion into fiat currency to minimize the risk. 

\vspace{.5ex}
An explanatory subgraph $G_{\zeta}^1$ is 
a factual explanatory subgraph generated by the 
explainer in~\cite{ying2019gnnexplainer}; 
and 
$G_{\zeta}^2$ is a counterfactual one generated by~\cite{lucic2022cf}. 
Comparing $G_{\zeta}^1$ with $G_{\zeta}^2$, 
we observe the following. 
(1) $G_{\zeta}^1$ is a small and concise explanation that 
only involves 
$v_t$ and most of its neighbors, 
which includes neighborhood features that preserve the output of 
a \gnn's output; nevertheless, it misses important 
nodes that have a higher influence on \gnn's output, 
that are not necessarily in $v_t$'s direct neighborhood. 
(2) Such nodes can be captured by a 
counterfactual explanatory subgraph, 
as depicted in $G_{\zeta}^2$. 
However, a larger fraction of $G$ is included 
to ensure that the ``removal'' of edges incurs enough 
impact to change the output of the \gnn, hence 
sacrificing ``conciseness'', conflicting with users' need on 
quick generation of small evidence which is ``faithful'' to the
original \gnn output~\cite{liu2021multi}. 

Moreover, choosing either alone can be biased to a ``one-sided'' explanation for the ``illicit'' IP address, grounded by one of the two 
money laundering patterns, but not both. 
\end{example}

Can we generate explanations that {\em simultaneously}
address multiple explainability metrics?
A quick solution is to compute subgraphs that  
optimize a weighted linear combination of all metrics~\cite{marler2010weighted,das1997closer}. 
However, this may lead to a marginally optimal answer over all measurements, overlooking other high-quality and diverse solutions, hence an overkill. 

\begin{example}
\label{exa-conflict}
Consider another two explanatory subgraphs: $G_{\zeta}^3$ is the explanation generated by~\cite{luo2020parameterized}, an explainer that optimizes both conciseness and factual measures~\cite{luo2020parameterized};
$G_{\zeta}^4$ is from an explainer that linearly combines factual and counterfactual measures into a single, bi-criteria objective function to be optimized~\cite{tan2022learning}. 
Such methods enforce to optimize potentially ``conflicting'' measures, 
or highly correlated ones, either may result in lower-quality solutions 
that are sensitive to data bias. 
For example, as fidelity$^-$ and conciseness both encourage 
smaller and faithful explanations, explanatory subgraphs obtained by ~\cite{luo2020parameterized} such as
$G_{\zeta}^3$ turns out to be relatively much smaller and less 
informative, hardly providing sufficient real-world evidence 
that can be used to interpret money laundering behaviors.  
On the other hand, explanations from~\cite{tan2022learning} 
such as $G_{\zeta}^4$ may easily be a one-sided, factual or counterfactual 
evidence. We found that such explanations capture only one type of money 
laundering scenario at best in most cases.  
\eat{
For example, $G_{\zeta}^3$ prioritizes the most concise explanation,
resulting in an explanatory subgraph with hardly any money laundering interpretability. 
Meanwhile, $G_{\zeta}^4$ captures only one type of money laundering scenario, called \textit{Spindle}~\cite{spindle} (defined later), while missing other high-quality explanations, thereby falling short in providing diverse money laundering interpretations that $v_t$ is involved.
}
\end{example}

Skyline queries~\cite{sky_opr, skylinequery3, skylinequery4, lin2006selecting, KungLP75} compute 
Pareto sets (``skylines'') that dominate the rest data points across a 
set of quality measures. 
Skylines generally offer better and more comprehensive solutions against 
the aforementioned alternatives \cite{das1997closer,sharma2022comprehensive}.
We advocate approaching \gnn explanation 
by generating a set of subgraphs 
that are Poreto sets over multiple user-defined 
explanatory measures, referred to as ``{\em skyline explanations}''. 

\begin{example}
\label{exa-skyline}
Consider a ``skyline query'' that seeks 
subgraphs addressing multiple explainability criteria, 
which returns a set of subgraphs $G_{\zeta}^5$, $G_{\zeta}^6$, and $G_{\zeta}^7$. These explanatory subgraphs are selected as a Pareto-optimal set across three explanatory measures: fidelity+, fidelity-, and conciseness. Each subgraph is high-quality, diverse, and non-dominated in at least one measure that is higher than others in the set. 
This result provides a more comprehensive and intuitive interpretation to 
explain ``why'' $v_t$ is identified as ``illicit'' by \gnn-based classification. 
Indeed, $G_{\zeta}^5$, $G_{\zeta}^6$, and $G_{\zeta}^7$ capture different money laundering scenarios: \textit{Peel Chain}~\cite{peel}, \textit{Spindle}~\cite{spindle}, and a combination, respectively.  
Therefore, such a skyline explanation supports law agencies by revealing a more comprehensive interpretation of the decision-making of \gnns on $v_t$. 
\end{example}

We advocate for developing a \gnn explainer that 
can efficiently generate skyline explanations for large-scale \gnn-based 
analysis. 
Such an explainer should:  
(1) generate skyline explanations for designated output of interest and 
any user-defined set of explanatory measures;  
and 
(2) generate a diversified set of skyline explanations upon request; 
and (3) ensure desirable guarantees in terms of Pareto-optimality. 
The need for 
skyline 
explanations are evident 
for trustworthy and 
multifaceted analysis and decision making, as observed in, \eg
drug repurposing~\cite{pushpakom2019drug}, cybersecurity analysis~\cite{cyber}, 
fraud detection~\cite{aml1},  
social recommendation~\cite{socialRecom}, 
among others, where 
\gnn output should be clarified 
from multiple, comprehensive aspects, rather than 
one-sided, biased perspectives. 

\stitle{Contributions}. 
We summarize our contributions as follows:

\sstab
(1) \emph{Formulation of Skyline Explanations.}
We introduce a class of skyline explanatory queries (\sgxq) 
that returns a skyline set of explanatory subgraphs optimizing multiple explainability measures. We formalize subgraph dominance and prove the computational hardness of query evaluation problem.

\sstab
(2) \emph{Algorithmic Solutions.}
We develop efficient algorithms for evaluating \sgxqs: an onion-peeling method with $(1+\epsilon)$-approximation, a diversification strategy to capture structural and embedding-level variety, and parallel algorithms that scale well to \eg billion-edge graphs.

\sstab
(3) \emph{Extensive Empirical Evaluation.}
Experiments on diverse real-world graphs show that our method generates more comprehensive explanations and scales effectively. For example, on \cora (citation network), it improves the Integrated Preference score of state-of-the-art explainers by $2.8\times$, on 
and achieves an $8.2\times$ speedup over \gnnexp\ on the million-edge \arxiv (citation network).

\stitle{Related Work}. We categorize related work into the following. 

\eetitle{Graph Neural Networks}.
GNNs have demonstrated themselves as powerful tools in performing various graph learning tasks. 
Recent studies proposed multiple variants of the GNNs, such as graph convolution networks (GCNs)~\cite{kipf2016semi}, graph attention networks (GATs)~\cite{gat}, Graph Isomorphism Networks (GINs)~\cite{gin}. These methods generally follow an information aggregation scheme where features of a node are obtained by aggregating features from its neighboring nodes.

\eetitle{Explanation of GNNs}. 
Several \gnn explanation approaches have been studied. 
(1) Learning-based methods aim to 
learn substructures of underlying graphs that 
contribute to the output of a \gnn. 
GNNExplainer~\cite{ying2019gnnexplainer} identifies subgraphs with node features that maximize the influence on the prediction by learning continuous soft masks for both adjacency matrix and feature matrix.
CF-GNNExplainer~\cite{lucic2022cf} 
learns the counterfactual subgraphs, which 
lead to significant changes of the 
output if removed from the graphs. 
PGExplainer~\cite{luo2020parameterized} parameterizes the learning process of mask matrix using a multi-layer perceptron.
GraphMask~\cite{schlichtkrull2020interpreting} learns 
to mask the edges through each layer of \gnns that 
leads to the most sensitive changes to their output. 
(2) Learning-based \gnn explainers require prior 
knowledge of model parameters and incur learning overhead for large graphs. Post-hoc approaches 
perform post-processing to directly compute subgraphs that optimize 
a fixed criteria. For example, ~\cite{yuan2021explainability} utilizes Monte-Carlo tree search to compute subgraphs that optimize a game-theory-inspired Shapley value. 
~\cite{zhang2022gstarx} follows a similar approach, yet optimizes HN values, a topology-aware variant of Shapley value. 
SAME~\cite{ye2023same} extends structure-aware Shapley explanations by using expansion-based Monte Carlo tree search to extract multi-grained connected substructures.
These methods are constrained to
pre-defined criterion, and cannot easily adapt to customizable, 
multiple explanability criteria. 
Beyond post-hoc explainers, GSAT~\cite{miao2022interpretable} uses stochastic attention within the information bottleneck framework to filter irrelevant components, while PGIB~\cite{seo2023interpretable} combines prototype learning with the bottleneck to highlight prototypical subgraphs.


Closer to our work are \gnn explainers that 
optimize pre-defined, multiple criteria. 
CF$^2$~\cite{tan2022learning} 
optimizes a linear function of factual and counterfactual measures and learns feature and edge masks as explanations. 
RoboGExp~\cite{robogexp} generates factual, counterfactual, and 
robust subgraphs, which remain unchanged 
upon a bounded number of edge modifications. 
GEAR~\cite{zhang2024gear} 
learns \gnn explainers by adjusting the gradients of multiple objectives geometrically during optimization. 
\moe\ ~\cite{liu2021multi} solves a bi-objective optimization problem to find Pareto optimal set over ``simulatability" (factual) and ``counterfactual relevance". 
Despite these methods generating explanations 
over multiple criteria, the overall 
goal remains pre-defined -- 
and not configurable 
as needed. In addition, \moe\ does not control the size of explanations, which may result in large number of explanations that are hard to inspect. 

\eat{
\eetitle{Skyline queries}.
Multi-objective search and skyline queries have been extensively 
studied \cite{skylinequery, skylinequery2, sky_opr}. 
These approaches compute 
Pareto optimal sets~\cite{hwang2012multiple, chircop2013constraint} or their approximate variants~\cite{websource, archiving} over data points 
and a set of optimization criteria. Notable strategies include~\cite{hwang2012multiple} that transform multiple objectives into a single-objective counterpart. 
Constraint-based methods such as~\cite{chircop2013constraint} initialize a set of anchor points that optimize each single 
measure, and bisect the straight lines between pairs of anchor points with a fixed vertical separation distance. 
This transforms bi-objective optimization into a series of single-objective counterparts. Solving each derives an approximation of the Pareto frontier. $\epsilon$-Pareto set~\cite{websource, archiving} has been widely recognized as a desirable approximation for the Pareto optimal set. While these algorithms cannot be directly applied 
to answer \sgxq, we introduce   
effective multi-objective optimization algorithms 
to generate explanatory subgraphs with provable quality guarantees 
in terms of $\epsilon$-Pareto approximation. 
}
\section{Graphs and GNN Explanation}
\label{sec-pre}




\stitle{Graphs.}
A 
graph $G = (V,E)$ has a set of nodes $V$ and a set of edges $E\subseteq V\times V$. 
Each node $v$ carries a tuple $T(v)$ of attributes and their values. 
The size of 
$G$, denoted as $|G|$, refers to the total number of its 
edges, \ie $|G|$ = $
|E|$.
Given a node $v$ in $G$, the {\em $L$-hop neighbors} of 
$v$, denoted as $N^L(v)$, refers to the set of all the nodes 
in the $L$-hop of $v$ in $G$. 
The {\em $L$-hop subgraph}  
of a set of nodes $V_s\subseteq V$, denoted as $G^L(V_s)$, 
is the subgraph 
induced by the node set $\bigcup_{v_s\in V_s}N^L(v_s)$. 

\stitle{Graph Neural Networks.}
\gnns~\cite{kipf2016semi,du2021multi,jiang2025ICML} comprise a well-established family of deep learning models tailored for analyzing graph-structured data. \gnns generally employ a multi-layer message-passing scheme as shown in Equation~\ref{eq-gnn1}.

\begin{small}
\begin{equation}
\vspace{-1ex}
\label{eq-gnn1}
    \mathbf{H}^{(l+1)} = \sigma(\widetilde{\mathbf{A}}\mathbf{H}^{(l)}\mathbf{W}^{(l)})    
\end{equation}
\vspace{-1ex}
\end{small}

$\mathbf{H}^{(l+1)}$ is the matrix of node representations at layer $l$, with $\mathbf{H}^{(0)}=\mathbf{X}$ being the input feature matrix. $\widetilde{\mathbf{A}}$ is a normalized adjacency matrix of an 
input graph $G$, which captures the topological feature of $G$.  
$\mathbf{W}^{(l)}$ is a learnable weight matrix at layer $l$ (a.k.a ``model weights''). $\sigma$ is an activation function. 

The {\em inference process} of a \gnn 
$\M$ with $L$ layers takes as input a graph $G$ = $(X,\widetilde{\mathbf{A}})$, 
and computes the embedding $\mathbf{H}^{(L)}_v$ for each 
node $v\in V$, 
by recursively applying the update 
function in Equation~\ref{eq-gnn1}. 
The final layer's output $\mathbf{H}^{(L)}$ (a.k.a ``output embeddings'') is used to 
generate a task-specific {\em output}, 
by applying a post-processing layer (\eg a softmax function). 
We denote the task-specific output 
as $\M(v, G)$, for the output of a \gnn $\M$ 
at a node $v\in V$. 

\eetitle{Fixed and Deterministic Inference}. 
We say that a \gnn $\M$ has a {\em fixed} inference 
process if its inference process is specified by fixed model parameters, number of layers, and message passing scheme. It has a {\em deterministic} 
inference process 
if $\M(\cdot)$ generates the same result for the same input. 
We consider \gnns with fixed, deterministic 
inference processes for consistent and robust performance in practice. 

\begin{table}
\caption{Summary of notations}
\begin{small}
\centering
\resizebox{\columnwidth}{!}{
\begin{tabular}
{c|c}
\textbf{Notation} & \textbf{Description} \\
\hline 
$G$=$(V, E)$ & a graph $G$ with node set $V$ and edge set $E$ \\ 
$\M$, $L$ & a \gnn model with number of layers $L$ \\
$\M(v,G)$ & task-specific output of $\M$ over $v\in V$ \\
$V_T$; $v_t$; $s$ & a set of (test) nodes; $v_t\in V_T$; a state $s$ (subgraph) \\
$\G_\zeta$; $G_\zeta$; $\zeta$ & a set of explanatory subgraphs; $G_\zeta\in \G_\zeta$; interpretable space  \\ 
$\Phi$; $\phi$ & a set of explainability measures; $\phi\in\Phi$ \\
$k$; $\alpha$; $\epsilon$ & size of skyline set; dominance score approximation; dominance factor \\
\hline
\end{tabular}
}
\end{small}
\label{tab-notations}
\vspace{-4ex}
\end{table}

\stitle{Node Classification.} 
Node classification is a fundamental task in graph analysis~\cite{nc}. A \gnn-based 
node classifier learns a \gnn $\M: (X,\mathbf{A}) \rightarrow \mathbf{Y}$ s.t. $\M(v)=y_v$ for $v \in V_{Tr} \subseteq V$, where $V_{Tr}$ is the training set of nodes with known (true) labels $Y_{Tr}$.  
The inference process of a trained 
\gnn $\M$ assigns the 
label for a test node $v_t\subseteq V_T$, 
which are derived from their computed  
embeddings.

\stitle{GNN Explainers and Measures}.
\label{exp_subg_def}
Given a \gnn $\M$ and an output $\M(v,G)$ to be explained, 
an {\em explanatory subgraph} $G_\zeta$ is 
an edge-induced, connected subgraph of $G$ 
with a non-empty edge set $E_\zeta\subseteq E$ that are responsible 
to clarify the occurrence of $\M(v,G)$. 
We call the set of 
all explanatory subgraphs as an 
{\em interpretable space}, denoted as $\zeta$. 

A \gnn~{\em explainer} is an algorithm that generates 
explanatory subgraphs in $\zeta$ for $\M(v,G)$. 
An {\em explainability measure} $\phi$ is a function: 
$G_\zeta$$\rightarrow$$\R$ that associates an explanatory subgraph with an explainability score. 
Given $G$ and $\M(v,G)$ to be explained,  
existing explainers typically 
solve a single-objective optimization problem: 
$G_\zeta^*$ = $\argmax_{G_\zeta\in\zeta}$ $\phi(G_\zeta)$.

\vspace{.5ex}
Several \gnn explainers and the explanability measures that they aim to optimize, are 
summarized in Table~\ref{tab-measures}.  
We summarize the main notations in Table~\ref{tab-notations}.

\stitle{Skyline Query}.
A skyline query is a multi-criteria database query that retrieves a set of {\em non-dominated} records from a dataset, based on a given set of attributes~\cite{papadias2003optimal,papadias2005progressive}. A record dominates another if it is at least as good in all dimensions and strictly better in at least one. 
For example, when searching for hotels, a skyline query would return those that are not worse than any other option in both price and distance to the city center, while being strictly better in at least one of them. 
The result forms the {\em Pareto frontier} of the dataset, representing the most interesting or optimal choices without requiring users to specify explicit weighting functions across multiple criteria. Skyline queries are widely used in decision making, recommender systems, and multi-objective optimization tasks.

\section{Skyline Explanations}
\label{sec:prob}


As aforementioned in~Example~\ref{exa-motivation},  
explanations that optimize a single explainability measure may not be 
comprehensive for users' interpretation 
preference. Meanwhile, 
a single explanation that optimizes multiple 
 measures may not exist, as two measures 
 may naturally ``conflict''. Thus, we 
pursue high-quality explanations for \sgxq in terms of 
multi-objective optimality measures.  
We next introduce our skyline explanation structure and the corresponding generation problem. 

\subsection{Skyline Explanatory Query} 
\label{sec:query}

We start with a class of 
explanatory queries. 
A {\em Skyline explanatory query}, denoted as \sgxq, 
has a form  
$\sgxq(G, \M, v_t,\Phi)$, 
where (1) $G$ = $(V,E)$ is an input graph, 
(2) $\M$ is a \gnn; 
(3) $v_t\subseteq V_T$ is a designated test node of interest, 
with an output $\M(v_t,G)$ to be explained; 
and (4) $\Phi$, the {\em measurement space}, refers to a set of normalized explainability 
measures to be {\em maximized}, each has a range $(0,1]$. 
For a measure to be better minimized (\eg $\conc$, $\fminus$ in Table \ref{tab-measures}), one can readily convert it to an inverse counterpart.
To characterize the 
query semantic, we 
specify explanatory subgraphs and 
their dominance relation. 


\eetitle{Explanatory Subgraphs}. 
Given a node of interest $v_t\in V_T$, 
a subgraph $G_\zeta$ of $G$ is an {\em explanatory 
subgraph} 
\wrt the output $\M(v_t,G)$, if 
it is {\em either} a factual {\em or} 
a counterfactual 
explanation. That is, 
$G_\zeta$ satisfies {\em one} of the two conditions: 
\tbi 
\item $\M(v_t,G)$ = $\M(v_t,G_\zeta)$; 
\item $\M(v_t,G)\neq$ $\M(v_t,G\setminus G_\zeta)$
\ei 

The interpretable space $\zeta$ \wrt 
a skyline explanatory query $\sgxq(G, \M, v_t,\Phi)$
contains all the 
explanatory subgraphs \wrt 
output $\M(v_t, G)$.

\eat{
\eetitle{Explainability Measures.}
We make cases for three widely used explainability measures: 
$\fplus$ measures the counterfactual property of explanatory subgraphs. Specifically, we exclude the edges of the explanatory subgraph from the original graph and conduct the \gnn\ inference to get a new prediction based on the obtained subgraph. If the difference between these two results is significant, it indicates a good counterfactual explanatory subgraph. 
Similarly, $\fminus$ measures the factual property, i.e., how similar the explanatory subgraph is compared to the original graph in terms of getting the same predictions. 
$\conc$ intuitively measures how compact is the explanatory subgraph, i.e., the size of the edges.
}

\eetitle{Dominance}. Given a measurement space $\Phi$ 
and an interpretable space $\zeta$, we say that
an explanatory subgraph $G_\zeta\in \zeta$ is {\em dominated} 
by another $G'_\zeta\in\zeta$, denoted as $G_\zeta \prec G'_\zeta$, 
if 
\tbi 
\item for each measure $\phi\in\Phi$, $\phi(G_\zeta)\leq\phi(G'_\zeta)$; and 
\item there exists a measure $\phi^*\in\Phi$, such that 
$\phi^*(G_\zeta)<\phi^*(G'_\zeta)$. 
\ei

\eetitle{Query Answers}. 
Given $\sgxq(G, \M, v_t,\Phi)$ with an interpretable space $\zeta$, 
a set of explanatory subgraphs $\G_\zeta \subseteq \zeta$ 
is a {\em skyline explanation}, if 
\tbi
\item there is no pair 
$\{G_1,G_2\} \subseteq \G_\zeta$   
such that $G_1 \prec G_2$ or $G_2 \prec G_1$;  
and 
\item for any other  
$G\in \zeta\setminus\G_\zeta$, 
and any $G'\in \G_\zeta$,  
$G\prec G'$.
\ei
That is, $\G_\zeta$ is a Pareto set~\cite{sharma2022comprehensive} of the interpretable space $\zeta$. 

A skyline explanation may contain an excessive number of 
 subgraphs that are too many to inspect. We 
 thus pose a pragmatic cardinality 
constraint $k$. We say that a skyline explanation 
is a {\em $k$-explanation}, if it contains at most 
$k$ explanatory subgraphs ($k\leq|\zeta|]$). 
A $k$-skyline query (denoted as $\sgxq^k$) admits 
only $k$-explanations as its query answers. 




\subsection{Quality of Skyline Explanations}
\label{sec:formulate}


A \gnn output may still have multiple $k$-explanations that can be too many for users to inspect. 
Moreover, one $k$-explanation dominating significantly less explanations than another in $\zeta$, 
may be treated ``unfairly'' as equally good. To mitigate 
such bias, we 
introduce a notion of dominance power. 

\vspace{.5ex}
Given an explanatory subgraph $G_\zeta\in\zeta$,  
the dominance set of $G_\zeta$, denoted as 
$\dset(G_\zeta)$, 
refers to the largest set 
$\{G'|G'\prec G_\zeta\}$, \ie the set of all 
the explanatory subgraphs that are dominated by $G_\zeta$ in $\zeta$. 
The {\em dominance power} of a $k$-explanation 
$\G_\zeta$  
is defined as 
\begin{small}
\vspace{-1ex}
\begin{equation}
    \dscore({\G_\zeta}) = \bigg|\bigcup_{G_\zeta\in\G_\zeta}\dset(G_\zeta)~\bigg|
\label{dsscore}
\end{equation}
\vspace{-1ex}
\end{small}
Note that $\dscore({\G_\zeta})\leq |\zeta|$ for any explanation $\G_\zeta \subseteq \zeta$.  

\stitle{Query Evaluation}. 
Given a skyline explanatory query $\sgxq^k$ = $(G,\M,v_t,\Phi)$, the query 
evaluation problem, denoted as $\eval(\sgxq^k)$, is to 
find a $k$-explanation $\G_\zeta^{k*}$,   
such that 
\begin{small}
\begin{equation}
\G_\zeta^{k*} = \argmax_{\G_\zeta\subseteq\zeta,|\G_\zeta|\leq k} \dscore(\G_\zeta) 
\end{equation}
\end{small}




\subsection{Computational Complexity}
\label{sec:complexity}

We next investigate the hardness of evaluating 
skyline exploratory queries. 
We start with a 
verification problem. 


\eetitle{Verification of Explanations}. 
Given a query $\sgxq^k$ = $(G,\M,v_t,\Phi)$, 
and a $k$-set of subgraphs $\G$ of $G$, 
the verification problem is 
to decide if for all $G_s\in\G$, $G_s$ is a factual or a counterfactual explanation 
of $\M(v_t,G)$.


\begin{theorem}
\label{thm-verify}
The verification problem for $\sgxq^k$ is in P. 
\end{theorem}

\begin{proofS}
Given an $\sgxq$ = $(G,\M,v_t,\Phi)$ and a set of subgraphs $\G$ of $G$, 
we provide a procedure, denoted as \verify,
that correctly determines if $\G$ 
is an explanation. The algorithm checks, for each $G_s\in\G$, if $G_s$ is a factual or a counterfactual explanation 
of $\M(v_t,G)$. It has been verified that this process can be performed by invoking a polynomial 
time inference process of $\M$~\cite{chen2020scalable,robogexp} for $v_t$ over $G_s$ (for testing factual explanation) 
and $G\setminus G_s$ (for testing counterfactual explanations), respectively.  
\end{proofS}

While it is tractable to verify explanations, 
the evaluation of $\sgxq^k$ is already nontrivial for $|\Phi|$ = $3$, even for a constrained case that $|\zeta|$ is a polynomial of $|G|$, \ie there are polynomially 
many connected subgraphs to 
be explored. 


\begin{theorem}
\label{thm-hardness-tractable}
$\eval(\sgxq^k)$ is already \NP-hard 
 even when $|\Phi|=3$ 
and $|\zeta|$ is polynomially bounded by $|G|$.
\end{theorem}

\begin{proofS}
It suffices to show that $\eval(\sgxq^k)$ is \NP-hard for an interpretable space $|\zeta|$ = $f(|G|)$ for 
some polynomial function $f$, and $|\Phi|=3$. We show this by 
constructing a polynomial-time reduction 
from $k$-representative skyline selection problem 
($k$-RSP). 
Given a set $S$ of $d$-dimensional data points, 
$k$-RSP is to compute a subset $S^*$ of $S$, 
such that (a) $S^*$ is a set of skyline points, $|S^*|=k$,
and (b) $S^*$ maximizes the dominance score.  
$k$-RSP is \NP-hard 
even for $3$-dimensional data points~\cite{lin2006selecting}. 

Given an instance of $k$-RSP with a set $S$ of $n$ 
$3$-dimensional data points, 
the reduction 
sets a graph $G$ as a two-level tree 
with a single root $v_t$, and $n$ 
distinct edges $(v_i,v_t)$, one 
for each data point $s_i\in S$. 
All the nodes in $G$ are assigned the same 
feature vector and the same label $l$. 
Duplicate graph $G$ as a training graph $G_T$ and train a 
vanilla \gcn $\M$ with $L$=$2$, with an output 
layer a trivial function that sets an output 
$\M(v_t, G)$ = $l$. As such, any 
sub-tree rooted at $v_t$ in $G$ is a factual 
explanatory subgraph $G_\epsilon$ for the output 
$\M(v_t, G)$ = $l$. For $\Phi$, we fix one 
of the measures $\phi_1\in\Phi$ to be ``conciseness'', 
hence enforcing the interpretable 
space to contain only 
singleton edges, \ie 
any explanatory subgraph $G_\epsilon'$ with more than one 
edges is strictly dominated by a single edge counterpart. 

We next \textit{break ties} of all size-$1$ 
explanatory subgraphs as follows. 
For each $3$-dimensional data point $s_i\in S$ 
as a triple $(d^1_i, d^2_i, d^3_i)$, 
recall that $s$ dominate $s'$ iff 
$s[i]\geq s'[i]$ for $i\in [1,3]$, and 
there is at least one dimension $j$ 
such that $s[j]>s'[j]$. 
Assume \kwlog that  
dimension $3$ is the ``determining''
dimension asserting the 
dominance of data points. 
Accordingly, we set 
$\phi_3$ as the determining 
dimension, and $\phi_2$ 
an order-preserving encoding of 
dimensions $1$ and $2$
that preserve original dominance, 
for each pair of data points in $S$. 
This obtains, for each $G_\epsilon$, 
a $3$-D measure $\Phi(G_\epsilon)$ 
accordingly. We can verify 
that $s$ dominates $s'$, 
iff $G_\epsilon$ =
$(v_s, v_t)$ dominates 
$G'_\epsilon$ = $(v_s',v_t)$. 
Given that $|\zeta|$ = $|G|$, 
the above construction is in polynomial time. 
Consistently, a $k$-representative 
set $S'$ for $k$-RSP induces an optimal $k$-explanation for $\M(v_t, G)$, 
and vice versa. The hardness 
of $\eval(\sgxq^k)$ hence follows.  
\end{proofS}
A straightforward 
evaluation of $\sgxq^k$ may explore 
the $L$-hop subgraph $G^L(\{v_t\})$ 
and initializes the interpretable space $\zeta$ as 
all {\em connected} subgraphs 
in $G^L(\{v_t\})$, by invoking 
subgraph enumeration algorithms~\cite{karakashian2013algorithm,Yang21}. 
It then 
enumerates $n$ 
size-$k$ subsets of $\zeta$ and finds 
an optimal $k$-explanation.  
Although this correctly finds  
optimal explanations, it 
is not practical for large $G$, as $n$ alone can be $2^{deg^L}$ 
and the Pareto sets 
inspection can take
$O(\binom{n}{k})$ time. 
We thus consider ``approximate 
query processing'' scheme for $\sgxq^k$, and present efficient 
algorithms that do not require such 
enumeration. 

\section{Generating Skyline Explanations}
\label{sec-alg}

\subsection{Approximating Skyline Explanations}
\label{sec-approx}

We introduce our first algorithm, which answers skyline 
explanatory queries with relative quality guarantee. To this end, we 
introduce a notion of 
$\epsilon$-explanation.

\stitle{$\epsilon$-explanations}.
Given explanatory measures $\Phi$ and an interpretable space $\zeta$, we say that
an explanatory subgraph $G_\zeta\in \zeta$ is {\em $\epsilon$-dominated} 
by another $G'_\zeta\in\zeta$, denoted as $G_\zeta \preceq_\epsilon G'_\zeta$, 
if 
\tbi 
\item for each measure $\phi\in\Phi$, $\phi(G_\zeta)\leq (1+\epsilon) \phi(G'_\zeta)$; and
\item there exists a measure $\phi^*\in\Phi$, such that 
$\phi^*(G_\zeta)\leq\phi^*(G'_\zeta)$. 
\ei 

Given  
$\sgxq^k$ = $(G,\M,v_t,\Phi)$, $\M$, and $v_t$, an explanation $\G_\epsilon\subseteq \zeta$ 
is an {\em $(\zeta, \epsilon)$-explanation} 
\wrt $G$, $\M$, and $v_t$, if (1) 
$|\G_\epsilon|\leq k$, and (2) for any explanatory subgraph $G_\zeta\in\zeta$, there is an explanatory 
subgraph $G'_\zeta\in\G_\epsilon$, 
such that  $G_\zeta \preceq_\epsilon G'_\zeta$.

In other words, a $(\zeta, \epsilon)$-explanation $\G_\epsilon$ 
approximates a $k$-explanation $\G_\zeta$ as its answer in 
the interpretable space $\zeta$. Indeed, 
(1) $\G_\epsilon$ has a bounded number $k$ of 
explanatory subgraphs as $\G_\zeta$; 
(2) $\G_\epsilon$ is, by definition, an {\em $\epsilon$-Pareto set}~\cite{schutze2021computing}  
of $\zeta$. In multi-objective decision making, 
$\epsilon$-Pareto sets have been verified as
a class of cost-effective, size-bounded approximation 
for Pareto optimal solutions. 

\stitle{$(\alpha, \epsilon)$-Approximations}. 
Given a $k$-skyline query $\sgxq^k$ $(G,\M,v_t,\Phi)$, 
and an interpretable space $\zeta$ \wrt $G$, $\M$, and $v_t$, 
let $\G_\zeta^*$ be the optimal $k$-explanation 
answer for $\sgxq^k$ in $\zeta$ (see \S~\ref{sec:formulate}). 
We say that an algorithm is an 
{\em $(\alpha, \epsilon)$-approximation} 
for the problem $\eval(\sgxq^k)$ \wrt $\zeta$, 
if it ensures the following: 
\tbi 
\item it correctly computes an  
$(\zeta, \epsilon)$-explanation $\G_\epsilon$; 
\item $\dscore(\G_\epsilon)\geq \alpha\dscore(\G_\zeta^*)$; and 
\item it takes time in $O(f(|\zeta|, |G|, \frac{1}{\epsilon}))$, where $f$ is a polynomial.  
\ei 

We present our main result below. 

\begin{theorem}
\label{thm-approx} 
There is a $(\frac{1}{4},\epsilon)$-approximation 
for $\eval(\sgxq^k)$ \wrt 
$\zeta'$, where $\zeta'$ is the set of explanatory 
subgraphs verified by the algorithm. The algorithm 
computes a $(\zeta', \epsilon)$-explanation 
in time $O(|\zeta'|(\log\frac{r_\Phi}{\epsilon})^{|\Phi|}+|\zeta'|L|G^L(v_t)|)$. 
\end{theorem}

Here (1) $r_\Phi$ = $\max\frac{\phi_u}{\phi_l}$,
for each measure $\phi\in\Phi$ with a range 
$[\phi_l, \phi_u]$; (2) $G^L(v_t)$ 
refers to the $L$-hop neighbor subgraph of node $v_t$, and (3) $L$ is the 
number of layers of the \gnn $\M$. 
Note that in practice, $|\Phi|$, $L$, and $\epsilon\in[0,1]$ 
are small constants, and $r_\Phi$ is often small. 

As a constructive proof of Theorem~\ref{thm-approx}, 
we introduce an approximation 
algorithm for $\eval(\sgxq^k)$. 

\stitle{Algorithm}. 
\label{sec-apxsx}
Our first algorithm, denoted as~\apxsx (illustrated as Algorithm~\ref{alg:apx}), takes advantage of a \textit{data locality} property: For 
a \gnn $\M$ with $L$ layers and any node $v_t$ in $G$, 
its inference only involves the nodes up to 
$L$-hop neighbors of $v$ via message passing, 
regardless of how large $G$ is. Hence, 
it suffices to explore and verify connected subgraphs 
in $G^L(V_T)$ (see \S~\ref{sec-pre}). 
In general, it interacts with three procedures: 

\sstab 
(1) a \kw{Generator}, which initializes and 
dynamically expands a potential  
interpretable space $\zeta'$, by 
generating a sequence 
of candidate explanatory subgraphs 
from $G^L(V_T)$; 

\sstab 
(2) a \kw{Verifier}, 
which asserts if an input candidate $G_s$ is 
an explanatory subgraph for $\sgxq^k$; 
and 

\sstab 
(3) an \kw{Updater}, 
that dynamically maintains a  
current size $k$ $(\zeta', \epsilon)$-
explanation $\G_\epsilon$ over 
verified candidates $\zeta'$, 
upon the arrival of verified 
explanatory subgraph in (2), 
along with other auxiliary 
data structures. 
The currently maintained explanation 
$\G_\epsilon$ is returned 
either upon termination (to be discussed), or upon 
an ad-hoc request at any time 
from the queryer.

\eetitle{Auxiliary structures}. 
 \apxsx coordinates the interaction of the \kw{Generator}, \kw{Verifier},  
and \kw{Updater} 
via a state graph (simply 
denoted as $\zeta'$). Each node (a ``state'') $s\in\zeta'$ 
records a candidate $G_s$ and its local information 
to be updated and used for evaluating $\sgxq^k$. 
There is a directed edge (a ``transition'')  
$t$ = $(s,s')$ in $\zeta'$ if $G_{s'}$ is obtained by 
applying a graph editing operator (\eg edge insertion, 
edge deletion) to $G_s$. 
A path $\rho$ in the state graph $\zeta'$ consists of 
a sequence of transitions that results in a 
candidate. 
In addition, each state $s$ is associated with 
(1) a score $\dscore(G_s)$, 
(2) a coordinate $\Phi(s)$, 
where each entry 
records an explainability measure $\phi(G_s)$ 
($\phi\in\Phi$); 
and (3) a variable-length bitvector $B(s)$, 
where an entry $B(s)[i]$ is $1$ if 
$G_i \preceq_\epsilon G_s$, 
and $0$ otherwise.  
The vector $B(s)$ keeps the 
$\epsilon$-dominance relation between $G_s$ 
and current candidates in $\zeta'$. 
Its $\dscore(G_s)$ score over $\zeta'$, 
can be readily counted 
as the number of ``1'' entries in $B(s)$.

\begin{algorithm}[tb!]
\renewcommand{\algorithmicrequire}{\textbf{Input:}}
\renewcommand{\algorithmicensure}{\textbf{Output:}}
\caption{\apxsx\ Algorithm}
    \begin{algorithmic}[1]
        \Require 
        a query $\sgxq^k$ = $(G,\M, v_t,\Phi)$; 
        a constant $\epsilon\in[0,1]$; 
        \Ensure 
        a $(\zeta',\epsilon)$-explanation $\G_\epsilon$. 
         \State set $\G_\epsilon$:=$\emptyset$; 
         \State identify edges for each hop: $\eset$:=$\{E_L, E_{L-1},\ldots,E_1\}$; \label{cd-onion}
        \For{$l$ = $L$ to $1$} \label{cd-op} 
            \State initializes state $s_{0}$:=$G^l(v_t)$. 
            \While{$E_l\neq \emptyset$}
                \For{$e\in E_l$} \label{cd-visit}
                    \State spawns a state $s$ with candidate $G_s$:=$G^l\backslash \{e\}$; \label{cd-spawn}
                    \State update $\zeta'$ with state $s$ and new transaction $t$;  
                    \If{$\veriF(s)$=False \& $\veriCF(s)$=False}
                    \label{cd-vrfy1}
                        \State continue;  
                    \EndIf \label{cd-vrfy2}
                    \State\label{cd-move}
                    $\G_\epsilon$:=\updatesx$(s, G_s,\zeta', \G_\epsilon)$; 
                    $E_l$:=$E_l\backslash\{e\}$; 
                    \label{cd-update}
                \EndFor
            \EndWhile 
        \EndFor\\
        \Return $\G_\epsilon$.    
    \end{algorithmic}
  \label{alg:apx}
\end{algorithm}

\stitle{``Onion Peeling''}. 
To reduce unnecessary verification costs, 
~\apxsx adopts a prioritized 
edge deletion strategy called ``onion peeling''. 
Given a node $v$ and its $L$-hop neighbor 
subgraph $G^L(v)$, it starts with 
a corresponding initial state $s_0$ that
iteratively removes  
edges from the ``outmost'' 
$L$-th hop ``inwards'' to $v$ (via \kw{Generator}). 
This spawns a set of new candidates to be 
verified (\kw{Verifier}), 
and 
maintained (\kw{Updater}). 

\vspace{.5ex}
This strategy enables several advantages. 
(1) Observe that 
$\M(G^L(v),v)$ = $\M(G,v)$ due to 
data locality. Intuitively, it 
is more likely to discover 
explanatory subgraphs earlier, by 
starting from candidates with small difference 
to $G^L(v)$, which is by itself 
a factual explanatory subgraph. 
(2) The strategy fully exploits 
the connectivity 
of $G^L(v)$ to ensure 
that the \kw{Generator} 
produces only 
connected candidates with $v$ 
included, over which 
$\dscore$, $\Phi$, and dominance 
are well defined. 
In addition, the 
process enables 
early detection and 
skipping of non-dominating 
candidates (see “Optimization”).

\eetitle{Outline}. Algorithm \apxsx 
dynamically maintains $\zeta'$ as the 
state graph. It first induces and verifies the $L$-hop neighbor subgraph $G^L(v)$, and initializes 
state node $s_0$ (w.r.t $G_{s_0}$) with $G^L(v)$ and 
local information. It then induces 
$L$-batches of edge sets $E_i$, $i\in[1,L]$, 
from $G^L(v)$ for onion peeling processing. 
For each ``layer'' (line~\ref{cd-op}) ($E_l, 1\leq l\leq L$), 
the \kw{Generator} procedure iteratively  
selects a next edge $e$ to be removed 
from the current layer and 
generates a new 
candidate $G_{s'}$ by removing $e$ from $G_s$, spawning a new state $s'$ in $\zeta'$ 
with a new transaction $t$ = $(s,s')$.
Following this procedure, we obtain a “stream” of states to be verified. 
Each candidate is then processed by the
\kw{Verifier} procedures, \veriF\ and \veriCF, 
to test if 
$G_s$ is factual or counterfactual, respectively (lines~\ref{cd-vrfy1}-\ref{cd-vrfy2}). 
If $G_s$ passes the test, 
the associated state $s\in \zeta'$ is 
processed by invoking the \kw{Updater} procedure 
\updatesx, in which the coordinator $\Phi(s)$, 
$(1+\epsilon)$-dominance relation (encoded in $B(s)$), and $\dscore(s)$ are incrementally updated (line~\ref{cd-compute}). 
~\updatesx then 
incrementally maintains the current 
explanation 
$\G_\epsilon$ with the newly verified 
explanatory subgraph $G_s$, following  
a replacement strategy (see Procedure~\updatesx). 
The processed edge $e$ is then 
removed from $E_l$ 
(line~\ref{cd-move}). 

\begin{figure}[tb!]
    \centering
\includegraphics[width=0.45\textwidth]{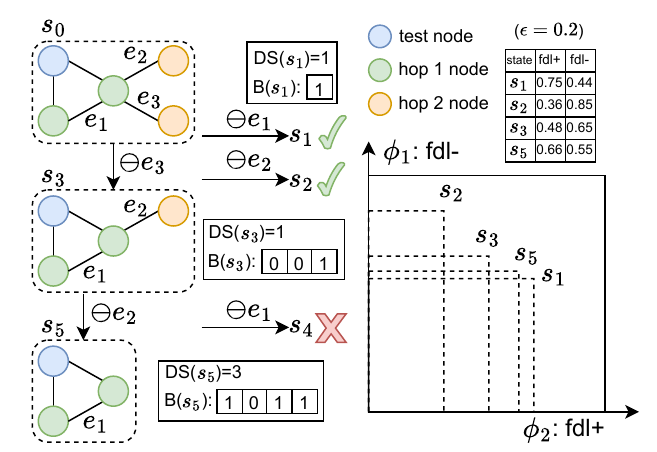}
    \caption{Illustration of Onion Peeling ($L$=$2$, $k$=$2$). $\zeta'$ = $\{s_1, s_2, s_3, s_5\}$. $(\zeta',\epsilon)$-explanation 
    $\G_\epsilon$=$\{s_2, s_5\}$ with $\dscore$ = $4$.}   
    \vspace{-3ex}
    \label{fig:running}
\end{figure}

\begin{example}
\label{exa-onion}
Consider a $\sgxq^2$ query in Figure~\ref{fig:running} 
that requests $2$-explanations with a measure space $\Phi$ 
that requires high $\fplus$ and $\fminus$. ~\apxsx starts the generation of 
subgraphs within the 2-hop subgraph $s_0$, by peeling single edges at hop $2$ 
of $v_t$, i.e., $e_1$, $e_2$, and $e_3$. This 
spawns three states $s_1$, $s_2$, and $s_3$ 
to be verified.  It chooses $s_3$, which leads to the new states $s_4$ and $e_5$,  
in response to the removal of $e_1$ and $e_2$, respectively. It continues to verify  
$s_4$ and $s_5$. As $s_4$ fails the verification (neither factual nor counterfactual), it continues to 
verify $s_5$ and obtains a current 
solution $\{s_3, s_5\}$. 
\end{example}

\begin{algorithm}[tb!]
\floatname{algorithm}{Procedure}
\renewcommand{\algorithmicrequire}{\textbf{Input:}}
\renewcommand{\algorithmicensure}{\textbf{Output:}}
\caption{Procedure \updatesx} 
    \begin{algorithmic}[1]
        \Require 
        state $s$, candidate $G_s$, state graph $\zeta'$, explanation $\G_\epsilon$;
        \Ensure updated 
        $(\zeta', \epsilon)$-explanation $\G_\epsilon$;  
        \State initializes state $s$ with $\dscore(s)$, $B(s)$:=$\emptyset$, $\Phi(G_s)$:=$\emptyset$;  \label{cd-compute}
        \State evaluates $\Phi(G_s)$; 
        \State incrementally determines $(1+\epsilon)$-dominance 
        of $G_s$; 
        \State updates $B(s)$ and $\dscore(s)$; 
        \If{$\{G_s\}$ is a new skyline explanation} 
        \label{cd-new} 
            \If{$|\G_\epsilon|$$<$$k$} 
                \State 
                $\G_\epsilon$ := $\G_\epsilon \cup \{G_s\}$; 
                \label{cd-no-full}
            \Else \label{cd-full1}
                 \State $\G_\epsilon$ := \swap$(\G_\epsilon, s)$; 
            \EndIf \label{cd-full2}
        \EndIf \label{cd-domd2}\\
\Return $\G_\epsilon$. 
    \end{algorithmic}
  \label{alg:updatesx}
\end{algorithm}

\eetitle{Procedure \updatesx}. 
For each new explanatory subgraph $G_s$ (at state $s$),  \updatesx 
updates its information 
by (1) computing coordinate $\Phi(G_s)$, 
(2) incrementally determines if $G_s$ is likely to join a 
skyline explanation in terms of 
$(1+\epsilon)$-dominance, \ie 
if for any verified 
explanatory subgraph $G_{s'}$ in $\zeta'$, 
$G_{s'}\preceq_\epsilon G_s$ (to be discussed). 
If so, and if the current explanation $\G_\epsilon$ 
has a size smaller than $k$, 
$G_s$ is directly added to $\G_\epsilon$. 
(line~\ref{cd-new}-~\ref{cd-no-full}). 
Otherwise, \updatesx performs a swap operator as follows: 
1) identify the skyline explanation $\sover\in \G_\epsilon$ that has the smallest $\dset(\sover)$; 
2) replace $\sover$ with $G_s$, only when 
such replacement makes the new explanation $\G_\epsilon'$ 
having a larger size 
$\dset(\G_\epsilon')$ increased by a factor of 
$\frac{1}{k}$ (line~\ref{cd-full1}-~\ref{cd-full2}). 
\eat{
We reduce the swap operation to the {\em MAX $k$-SET COVERAGE} problem discussed in~\cite{onlinemkc}. Specifically, swapping the new coming $G_\zeta$ only when the corresponding new $\gset^{k*}_{\zeta}$ is $(1+\frac{1}{k})$ large than the previous one. This will ensure a $\frac{1}{4}$-approximation ratio as proved in~\cite{onlinemkc}.
}

\eetitle{Update Dominance Relations}. 
Procedure \updatesx~maintains a set 
of skyline explanations $\G_s$   (not shown) in terms of $(1+\epsilon)$-dominance. 
The set $\G_s$ is efficiently derived 
from the bitvectors $B(s)$ 
of the states in $\zeta'$. 
The latter compactly encodes a lattice structure of 
$(1+\epsilon)$ dominance 
as a directed acyclic graph; 
and $\G_s$ refers to the states 
with no ``parent'' in the lattice, 
\ie having an explanatory subgraph 
that is not $(1+\epsilon)$-dominated by 
any others so far. 
Details are in Appendix. 

\begin{example}
\label{exa-replacement}
Recall Example~\ref{exa-onion} where the explanation space $\zeta'$ includes $\{s_1, s_2, s_3, s_5\}$.  $(1+\epsilon)$-dominance 
 is tracked by a dynamically maintained scoring table (top-right of Figure~\ref{fig:running}).  As a sequence of 
states $s_1$, $s_2$, $s_3$, $s_5$ is generated, the first two states 
are verified to form a Pareto set, 
and are added to the explanation $\{s_1, s_2\}$. 
Upon the arrival of $s_3$, since it introduces an improvement less than a factor of $1+\frac{1}{k}$ = $\frac{3}{2}$,  $\updatesx$ skips $s_3$. As $s_5$ $(1+\epsilon)$-dominates $s_1$ and $s_3$, it replaces $s_1$ with $s_5$ and 
updates the explanation to be $\{s_2, s_5\}$. 
\end{example}

\vspace{-1ex}

\stitle{Explainability}. Algorithm~\apxsx terminates 
as it constantly removes edges 
from $G^L(v_t)$ and verifies a finite number of candidates.  
We also show its quality guarantee, as stated below. 

\begin{lemma}
\label{lm-epsilon} 
Given a constant $\epsilon$, \apxsx 
correctly computes a $(\zeta', \epsilon)$-explanation of size $k$ defined on 
the interpretation space $\zeta'$, 
which contains all verified candidates. 
\end{lemma}

\begin{proofS}
We show the above result with a reduction 
to the multi-objective shortest path 
problem (\kw{MOSP})~\cite{tsaggouris2009multiobjective}. 
Given an edge-weighted graph $G_w$, 
where each edge carries a $d$-dimensional 
attribute vector $e_w.c$, it computes a 
Pareto set of 
paths from a start node $u$. 
The cost of a path $\rho_w$ in $G_w$ is 
defined as $\rho_w.c$ = $\sum_{e_w\in\rho_w}$ $e_w.c$. 
The dominance relation between two paths 
is determined by 
the dominance relation of their cost vector. 
Our reduction (1) constructs $G_w$ 
as the running graph $\zeta'$ with 
$n$ verified states and 
transitions; and 
(2) for each edge $(s, s')$, 
sets an edge weight as 
$e_w$ = $\Phi(s) - \Phi(s')$.  
Given a solution $\Pi_w$ 
of the above instance of \kw{MOSP}, 
for each path $\rho_w\in\Pi$, 
we set a corresponding path $\rho$ 
in $\zeta'$ that ends 
at a state $\rho_s$
, 
and adds it into $\G_\epsilon$.  
We can verify that 
$\Pi_w$ is an 
$\epsilon$-Pareto set 
of paths $\Pi_w$ in $G_w$, if and only 
if $\G_\epsilon$ is an $(\zeta', \epsilon)$-explanation 
of $\zeta'$. 
We show that~\apxsx 
performs a simpler process of 
the algorithm in~\cite{tsaggouris2009multiobjective}, 
which 
ensures to generate $\G_\epsilon$ as a 
$(\zeta', \epsilon)$-explanation. 
\end{proofS}

\begin{lemma}
\label{lm-alpha} 
\apxsx computes a $(\zeta', \epsilon)$-explanation 
$\G_\epsilon$ that ensures 
$\dscore(\G_\epsilon)\geq \frac{1}{4}\dscore(\G^*_\epsilon)$, 
where $\G^*_\epsilon$ 
is the size $k$ $(\zeta', \epsilon)$-explanation with maximum 
dominance power $\dscore(\G^*_\epsilon)$. 
\end{lemma}

\begin{proofS}
Consider procedure~\updatesx 
upon the arrival, at any time, 
of a new verified candidate $G_s$. (1) The above results 
clearly hold when $|\zeta'|\leq k$ 
or $|\G_\epsilon|\leq k$, 
as $\G_\epsilon$ is the only 
$(\zeta', \epsilon)$-explanation 
so far. (2) When $|\G_\epsilon| = k$, we reduce the approximate  
evaluation of $\sgxq^k$ 
to an instance of the 
{\em online MAX $k$-SET
} problem~\cite{onlinemkc}. 
The problem maintains 
a size-$k$ set cover with 
maximized weights. 
We show that \updatesx 
adopts a greedy 
replacement policy by 
replacing a candidate in $\G_\epsilon$ with the new 
candidate $G_s$ only when 
this leads to a $(1+\frac{1}{k})$ 
factor improvement 
for $\dscore$. This ensures a $\frac{1}{4}$-approximation ratio~\cite{onlinemkc}. 
Following the replacement 
policy consistently, 
~\updatesx ensures 
a $\frac{1}{4}$ approximation 
ratio for $(\zeta', \epsilon)$-explanations in $\zeta'$.  
\end{proofS}

\stitle{Time Cost}\label{op-time-cost}. 
As Algorithm~\apxsx processes candidate subgraphs in an online fashion from a stream, we analyze its time complexity in an output-sensitive manner in terms of $\zeta'$ at termination.  
It evaluates $|\zeta'|$ candidate subgraphs. For each candidate $G_s$, the \kw{Verifier} module (procedures~\veriF\ and \veriCF) checks its validity for \sgxq via two forward passes of the \gnn\ model $\mathcal{M}$, incurring a cost of $O(L|G^L(v_t)|)$ per candidate, assuming a small number of node features (cf. Lemma~\ref{thm-verify},~\cite{chen2020scalable,robogexp}). Hence, the total verification cost is $O(|\zeta'|L|G^L(v_t)|)$. Verified candidates are passed to the \kw{Updater} module ($\updatesx$), which maintains a $(1+\epsilon)$-approximate Pareto set in $\zeta'$ by solving a multi-objective dominance update over the state space. This takes at most $O\left(\prod_{i=1}^{|\Phi|} \left(\left\lfloor \log_{1+\epsilon} \frac{\phi_u}{\phi_l} \right\rfloor + 1 \right)\right)$ time per update, where $\phi_l$ and $\phi_u$ are the minimum and maximum values observed for each metric $\phi \in \Phi$. Approximating $\log_{1+\epsilon} \approx \frac{1}{\epsilon}$ for small $\epsilon$, the total maintenance cost becomes $O\left(|\zeta'|\cdot \left(\frac{\log r_\Phi}{\epsilon}\right)^{|\Phi|}\right)$, where $r_\Phi = \frac{\phi_u}{\phi_l}$. Therefore, the total time cost of \apxsx is in $O\left(|\zeta'|\cdot \left(\left(\frac{\log r_\Phi}{\epsilon}\right)^{|\Phi|} + L|G^L(v_t)|\right)\right)$.

Putting the above analysis together, 
Theorem~\ref{thm-approx} follows. 

\stitle{Optimization.}\label{Optimization} 
To further reduce verification cost, 
\apxsx uses two optimization strategies, 
as outlined below. 

\eetitle{Edge Prioritization}\label{edge-prior}. 
\apxsx adopts an edge prioritization 
heuristic to favor 
promising candidates 
with small loss of 
$\dscore$ and that are more likely to 
be a skyline explanation (\apxsx, line~\ref{cd-spawn}). It ranks each transaction $t$=$(s,s')$ 
in $\zeta$ based on a loss estimation of $\dscore$ 
by estimating $\Phi(s')$. 
The cost vector of each transaction is aggregated to an average weight based on each dimension, \ie $w(t)$ = $\frac{1}{|\Phi|}$ $\sum_{i=1}^{|\Phi|} (\phi_i(s)-\hat{\phi_i(s')})$. 
The candidates from spawned states with the smallest loss of 
$\dscore$ are preferred. This helps 
early convergence of high-quality explanations, 
and also promotes early detection of 
non-dominating candidates (see ``early pruning''). 

\eat{
Since we aim to minimize the transaction cost, the prioritized transaction $t^*$ has the minimum \as\ value. 
We delete $e^*=t^{*}.e$ to transact from $G^L$ to $G^L\backslash\{e^*\}$. Then start from $G^L\backslash\{e^*\}$, we rank the remaining edges $E_L\backslash\{e^*\}$ based on the transactions from $G^L\backslash\{e^*\}$ to $G^L\backslash\{e^*, e\}$. Inherently, we obtain the {\em new} prioritized transaction ${t^*}'$ and continue the process until the last edge from $E_L$ is deleted.
}

\eetitle{Early Pruning}. 
\apxsx also exploits a {\em monotonicity property} of 
measures $\phi\in 
\Phi$ to early determine the non-$\epsilon$-dominance 
of an explanatory subgraph.
Given $\zeta'$ and 
a measure $\phi\in\Phi$, 
we say $\phi$ is {\em monotonic} \wrt a path $\rho$ in $\zeta'$,  
if for a state $s$ with candidate $G_s$ and 
another state $s'$ with a subgraph $G_{s'}$ of $G_s$ 
on the same path $\rho$,  
${\phi_l(G_s)}\geq\frac{\phi_u(G_{s'})}{1+\epsilon}$, 
where $\phi_l(G_s)$ (resp. $\phi_u(G_{s'})$) is a lower bound estimation 
of $\phi(G_s)$ (resp. upper bound estimation of 
$\phi(G_{s'})$), \ie $\phi_l\leq \phi_l(G_s)\leq \phi(G_s)$ (resp. $\phi(G_{s'}) \leq \phi_u(G_{s'})\leq \phi_u$). 
By definition, for any $\phi$ with monotonicity property, we have $\phi(G_{s'})\leq \phi_u(G_{s'})\leq (1+\epsilon)\phi_l(G_s)\leq (1+\epsilon)\phi(G_s)$, hence $G_{s'}\preceq_\epsilon G_s$, 
and any such subgraphs $G_{s'}$ of $G_s$ can be safely 
pruned due to non-$(1+\epsilon)$ dominance determined by 
$\phi$ alone. 
Explainability measures such as density, 
total influence, conciseness, 
and diversity of embeddings, are likely to be  
monotonic. Hence, the onion peeling strategy 
enables early pruning by exploiting 
the (estimated) ranges of such measures. 
Note that the above property is checkable  
in $O(|\zeta'|)$ time.

\subsection{Diversified Skyline Explanations}
\label{sec-divsx}

\eat{
}

Skyline explanation may still contain explanatory subgraphs having highly ``similar''
or overlapped nodes. This may lead to 
redundant and 
biased explanations.  
In practice, diversified explanations 
often clarify the model output 
more comprehensively~\cite{gong2019diversity, mothilal2020explaining}. 
We next investigate a diversified scheme for 
 $\sgxq^k$ evaluation, in terms of coverage 
and the difference between node representations. 


\stitle{Diversification of Explanatory Query Evaluation}. 
Given a query $\sgxq^k$ = $(G,\M,v_t,k,\Phi)$, the diversified 
evaluation of $\diveval(\sgxq^k)$, is to 
find a $(\zeta',\epsilon)$-explanation $\G_\epsilon$, s.t.
\begin{small}
\begin{equation}
\G_\epsilon^* = \argmax_{\G\subseteq\zeta,|\G_\epsilon|\leq k} \divs(\G_\epsilon)
\end{equation}
\end{small}

where $\divs(\G_\epsilon)$ is a {\em diversification 
function} defined on $\G_\epsilon$, to quantify 
its overall diversity,  
%
and is defined as
\begin{small}
\begin{equation}
{
    \divs(\G_\epsilon) = \alpha \cdot \ncs(\G_\epsilon) + (1-\alpha)\cdot 
    \sum_{G_s,G_{s'}\in \G_\epsilon}\cd(G_s, G_{s'})
    \label{obj_func}
    }
\end{equation}
\end{small}
here $\ncs$, a {\em node coverage} measure, aggregates
the node coverage of explanatory subgraphs in $\G_\epsilon$:
\begin{small}
\begin{equation}
    \ncs(\G_\epsilon) = \frac{|\bigcup_{G_s\in\G_\epsilon} V_{G_s}|}{|V_{G^L}|}; 
\end{equation}
\end{small}
and $\cd$, an {\em accumulated difference} measure, aggregates the node difference 
between two subgraphs in terms of Cosine distances of their embeddings: 
\begin{small}
\begin{equation}
    \cd(G_s, G_{s'}) =  1 - \frac{\mathbf{x}_{G_s}\cdot \mathbf{x}_{G_{s'}}}{||\mathbf{x}_{G_s}||_2\cdot ||\mathbf{x}_{G_{s'}}||_2}
\end{equation}
\end{small}
where $\mathbf{x}_{G_s}$ is the embedding of $G_s$ obtained by node embedding 
learning such as Node2Vec~\cite{grover2016node2vec}. The two 
terms are balanced by a constant $\alpha$.

\stitle{Diversification Algorithm}. 
We next outline our diversified evaluation 
algorithm, denoted as \divsx (pseudo-code shown in Appendix). 
It follows~\apxsx and adopts onion peeling 
strategy to generate subgraphs in a stream. 
The difference is that when computing the $k$-skyline explanation, 
it includes a new replacement 
strategy when the marginal gain for 
the diversification function $\divs(\cdot)$ 
is at least a factor of the score over the current solution $\divs(\G_\epsilon)$. 
 \divsx  terminates when a first $k$-skyline explanation is found. 

\eetitle{Procedure \updatedivsx}. 
For each new candidate $G_s$ (state $s$),  \updatedivsx 
first initializes and updates $G_s$ 
by 
(1) computing its coordinates $\Phi(G_s)$,  
(2) incrementally determines if $G_s$ is a 
skyline explanation in terms of 
$(1+\epsilon)$-dominance, \ie 
if for any verified 
explanatory subgraph $G_{s'}$ in $\zeta'$, 
$G_{s'}\preceq_\epsilon G_s$. 
If so, 
1) check if the current explanation $\G_\epsilon$ 
has a size smaller than $k$; and 
2) the marginal gain of $G_s$ is bigger than $\frac{(1+\epsilon)/2-\divs(\G_\epsilon)}{k-|\G_\epsilon|}$. 
If 
$G_s$ satisfies both conditions, \updatedivsx adds it in the $k$-skyline explanation $\G_\epsilon$. 


\stitle{Quality guarantee}. 
Adding diversification 
still permits an approximation scheme with relative 
quality guarantees, 
with the same worst-case time cost. 

\begin{lemma}
Given a constant $\epsilon$, \divsx 
correctly computes a $(\zeta', \epsilon)$-explanation of size $k$ defined on 
the interpretation domain $\zeta'$, 
which contains all verified candidates. 
\end{lemma}

The proof is similar to Lemma 4, therefore we omit it here.

\begin{theorem}
\divsx computes a $(\zeta', \epsilon)$-explanation 
$\G_\epsilon$ that ensures 
$\divs(\G_\epsilon)\geq (\frac{1}{2}-\epsilon)\divs(\G^*_\epsilon)$, in time 
$O(|\zeta'|(\log\frac{r_\Phi}{\epsilon})^{|\Phi|}+|\zeta'|L|G^L(v_t)|)$; 
where $\G^*_\epsilon$ is the size-$k$ $(\zeta', \epsilon)$-explanation over $\zeta'$ with optimal diversity $\divs(\G^*_\epsilon)$. 
\end{theorem}

We show the above result by reducing the 
diversified evaluation of $\sgxq^k$ 
to an instance of the Streaming Submodular Maximization problem~\cite{ssm}. 
The problem maintains  
a size-$k$ set that optimizes a submodular function
over a stream of data objects. 
We show that \divsx adopts a consistent increment policy that ensures a ($\frac{1}{2}-\epsilon$)-approximation as in~\cite{ssm}. 
We present the detailed proof in Appendix.

\section{Parallel Skyline Explanatory Querying}\label{sec-parallel}

We next scale our approximate algorithms to query workloads. 
It is desirable to parallelize the evaluation of a batch of queries 
involving a set of targeted nodes $V_T$ to multiple threads. 
For example, a server may receive 
a set of \sgxq queries requesting fast explanations over a set $V_T$ of newly 
detected illicit IP addresses from multiple analysts. 

A random distribution of queries across threads may lead to redundant verification or ``stranglers'', due to the redundant computation of
overlapped neighbors of the nodes in $V_T$.
We thus introduce a parallel 
algorithm with a workload clustering strategy to maximize the parallelism. 

\eat{
\begin{figure}[tb!]
    \centering
\includegraphics[width=0.45\textwidth]{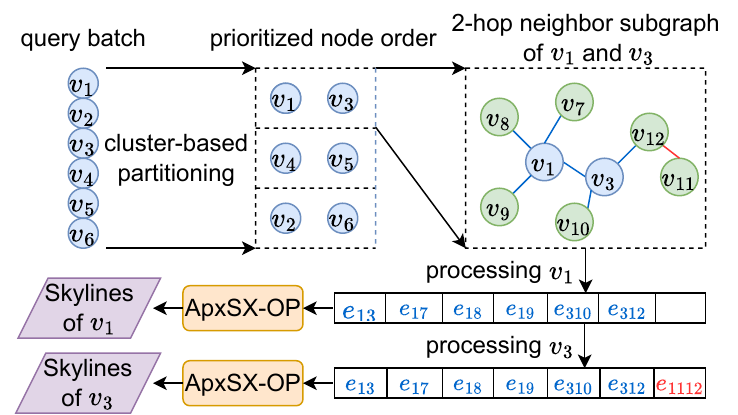}
    \caption{Parallelization with Clustering and Node Prioritization.}   
     \vspace{-3ex}
    \label{fig:para-frame}
\end{figure}
}

\stitle{Parallel Algorithm}. The parallel algorithm, denoted as \parasx, works with $m$ threads to evaluate a query set $Q$ with $n$ $\sgxq$ queries, 
involving a test node set $V_T$ 
(
pseudocode given in Appendix). (1) \textit{Inter-Batch parallelization}: It first partitions $Q$ into $N$ small batches to be evaluated in parallel,  guided by a clustering strategy 
with a goal that for any pair of target nodes $v_t$ and $v_t'$ from two different batches, 
their $L$-hop neighbors have minimized overlap (see ``Workload Clustering''). 
(2) \textit{Batch Assignment}. 
\parasx then solves a makespan minimization problem to assign the $N$ batches to 
$m$ threads, with a dynamic scheduling strategy that prioritizes the processing of 
queries with ``influencing'' output nodes that have highest common $L$-hop neighbors for those in the same thread. This is to maximize the shared computation for edge peeling and verification. 
(3) \textit{Inter-query Pipelining}: 
Each local thread then invokes \apxsx($v_t$) to generate local 
$k$-explanations for each assigned node $v_t$. All the processing 
follows its local ranks of highly influential nodes, and automatically 
skips processing a node, if its information 
has been updated via peeling operations earlier by 
previous nodes. This processing strategy is 
enabled by 
maintaining a shared data structure that records nodes 
and their $L$-hop neighboring edges within each thread. 
The skyline computation can be further parallelized by parallel skyline query methods~\cite{wu2006parallelizing, park2013parallel}.

\begin{figure*}[tb!]
    \centering
    \begin{subfigure}[b]{0.24\linewidth}
    \includegraphics[width=\linewidth]{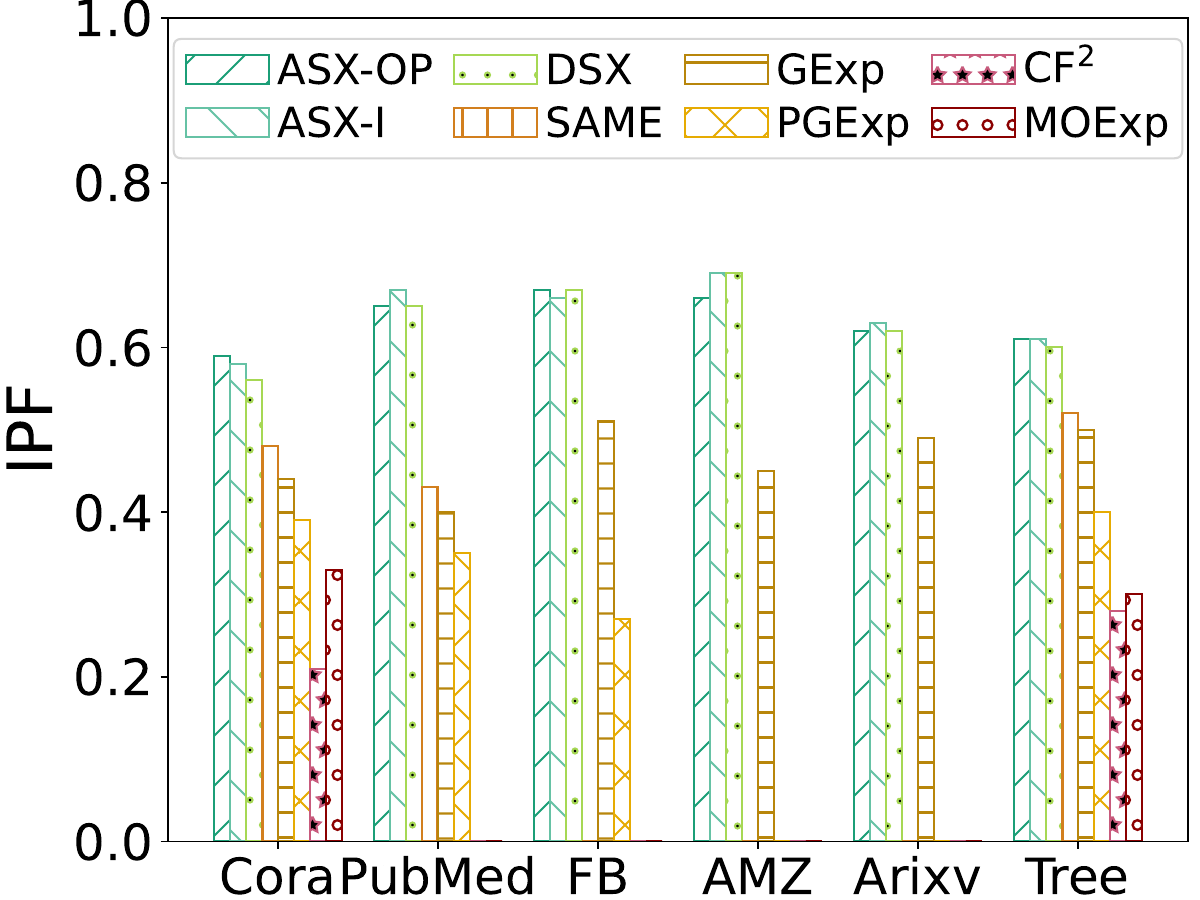}
        \caption{IPF score}
        \label{fig:ipf_effect}
    \end{subfigure}
    \hfill
    \begin{subfigure}[b]{0.24\linewidth}
        \includegraphics[width=\linewidth]{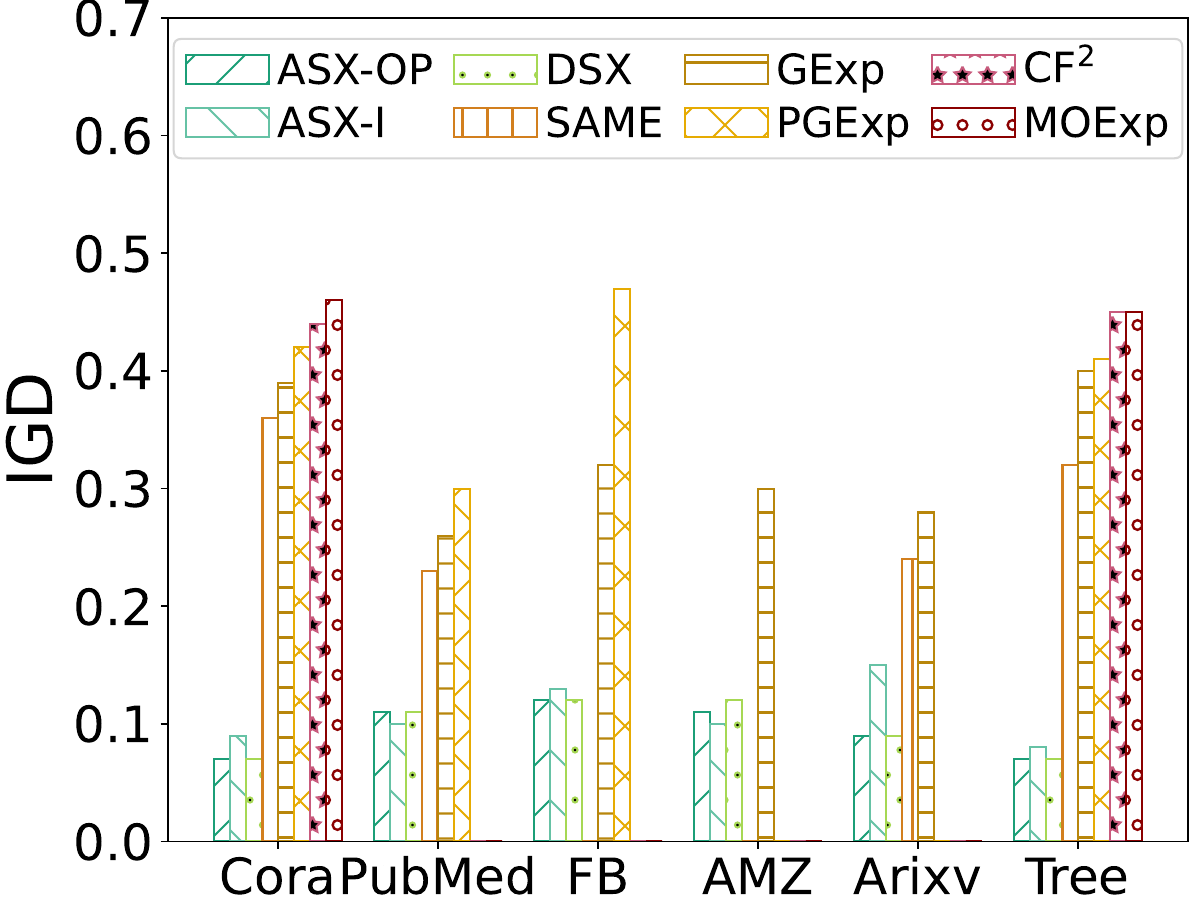}
        \caption{IGD score}
        \label{fig:igd_effect}
    \end{subfigure}
    \hfill
    \begin{subfigure}[b]{0.18\linewidth}
        \includegraphics[width=\linewidth]{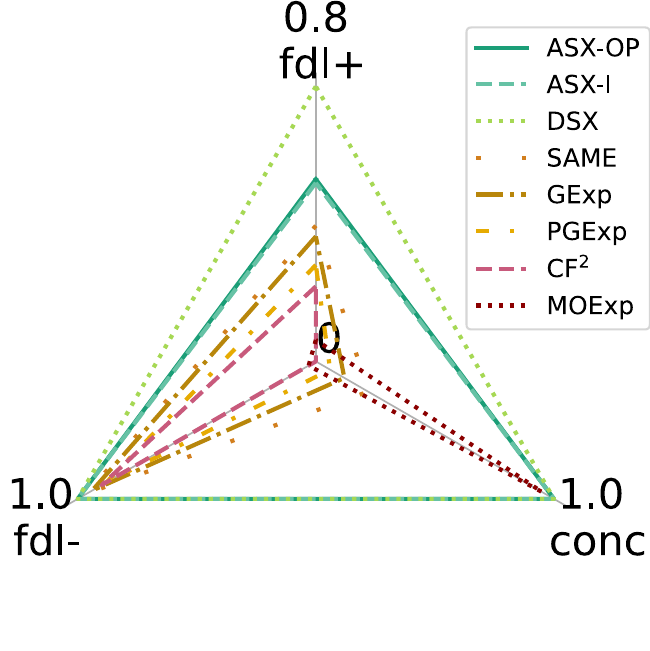}
        \caption{MS score (\cora)}
        \label{fig:radar1}
    \end{subfigure}
    \hfill
    \begin{subfigure}[b]{0.32\linewidth}
\includegraphics[width=\linewidth]{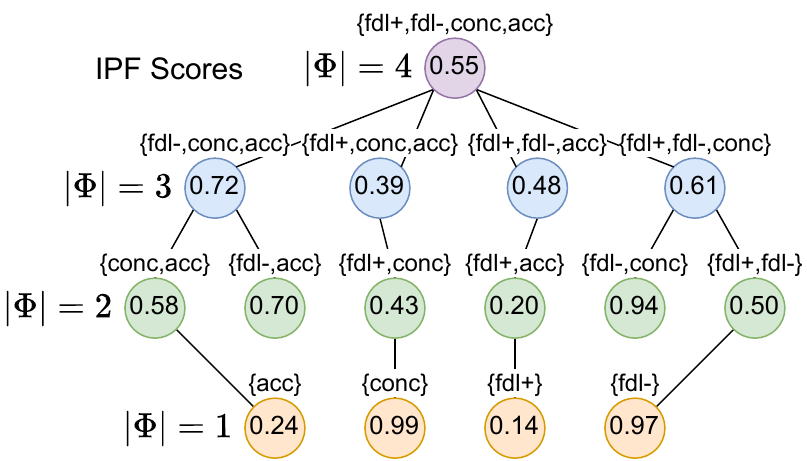}
        \caption{IPF score over subsets of $\Phi$  (\treecycle)}
        \label{fig:radar2}
    \end{subfigure}
    \caption{Effectiveness of our skyline explanation vs. baseline explainers (\gnnexp, \pgexp, \cff, \moe, \kw{SAME}).}
    \label{fig:overall_effect}
    \vspace{-3ex}
\end{figure*} 

\stitle{Parallelization Strategies}. \parasx employs two strategies.
(1) Workload Clustering. We partition query workloads by solving an $N$-clustering of $Q$ that maximizes intra-cluster Jaccard similarity of $L$-hop neighbors. This groups test nodes with overlapping neighborhoods, enabling efficient batched processing.
(2) Influential Node Prioritization. We apply $k$-core decomposition to rank nodes by structural centrality. Test nodes with higher core numbers are processed first, ensuring that edge information from influential nodes is shared early across threads for maximal reuse.

\begin{table}[tb!]
\caption{Statistics of datasets}
\label{tab:dataset}
\resizebox{\columnwidth}{!}{
\begin{tabular}{c|cccc}
dataset & \# nodes & \# edges & \# node features & \# class labels \\ 
\hline
\cora~\cite{cora} & 2,708 & 10,556 & 1,433 & 7 \\
\pubmed~\cite{pubmed} & 19,717 & 88,648 & 500 & 3 \\
\facebook~\cite{facebook} & 22,470 & 342,004 & 128 & 4 \\ 
\amazoncomputer~\cite{shchur2018pitfalls} & 13,752 & 491,722 & 767 & 10 \\
\arxiv~\cite{hu2020open} & 169,343 & 1,166,243 & 128 & 40 \\
\bahouse~\cite{ying2019gnnexplainer} & 1,010,000,000 & 6,027,852,000 & 1 
& 4 \\
\treecycle~\cite{ying2019gnnexplainer} & 991 & 2140 & 1 & 2 \\
\hline
\end{tabular}
}
 \vspace{-2ex}
\end{table}

\section{Experimental Study}
\label{sec-exp}

We conduct experiments to evaluate the effectiveness, efficiency, and scalability of our solutions. Our algorithms are implemented in Python 3.10.14 by PyTorch-Geometric framework. 
All experiments are conducted on a Linux system equipped with AMD Ryzen 9 5950X CPU, an NVIDIA GeForce RTX 3090, and 32 GB  of RAM. 
{\bf Our code and data are made available at~\cite{code_base}}.

\subsection{Experimental Setup}

\vspace{-1ex}
\stitle{Datasets.} Table~\ref{tab:dataset} summarizes the datasets.
(1) \cora~\cite{cora} and \pubmed~\cite{pubmed} are citation networks where nodes (papers) are classified by category using keyword-based features.
(2) \facebook\ (\kw{FB})~\cite{facebook} contains verified Facebook pages (nodes) linked by mutual likes, with features from page descriptions; the task is multi-class page classification.
(3) \amazoncomputer\ (\kw{AMZ})~\cite{shchur2018pitfalls} is a co-purchase network of computer products with review-based features; the task is product category classification.
(4) \arxiv\ (\kw{arxiv})~\cite{hu2020open} is a citation network of CS papers with 128-d word-embedding features, classified by subject area.
(5) \bahouse (\kw{BA})~\cite{ying2019gnnexplainer} is a synthetic BA graph with house motifs; house nodes serve as human-crafted ground-truth explanations.
(6) \treecycle (\kw{Tree})~\cite{ying2019gnnexplainer} is a synthetic tree graph with cycle motifs; cycle nodes serve as human-crafted ground-truth explanation~\cite{yuan2022explainability}.

\stitle{GNN Classifiers.} We employ three classes of mainstream \gnns:  
(1) \emph{Graph convolutional network} (\gcn)~\cite{kipf2016semi}, one of the classic message-passing \gnns;  
(2) \emph{Graph attention networks} (\gat)~\cite{gat} leverage  attention mechanisms to dynamically weigh the importance of a node’s neighbors during inference; and  
(3) \emph{Graph isomorphism networks} (\gin)~\cite{gin} with enhanced expressive power up to the Weisfeiler-Lehman graph isomorphism tests. 

\stitle{GNN Explainers.} We have implemented 
the following. 

\sstab 
(1) Our skyline exploratory query evaluation 
methods include 
two approximations 
\apxsx (\S~\ref{sec-apxsx}) 
and the diversification algorithm \divsx (\S~\ref{sec-divsx}). 
To evaluate the benefit of onion peeling strategy, 
we also implemented~\apxsxi, a variant of \apxsx 
that follows an ``edge insertion'' strategy to grow 
candidates up to its $L$-hop neighbor subgraph. 
(2) Parallel algorithm \parasx, and 
its variants \parasxn and \parasxeis. 
\parasxn is a naïve parallel algorithm that randomly assigns query loads to each thread
without any load balancing strategy.
\parasxeis is a parallel algorithm that only adopts asynchronous influential node prioritization with a global data structure 
 and randomly assigns query loads to each thread.

We compare against several state-of-the-art \gnn explainers.
(1) GNNExplainer (\gnnexp) learns edge and feature masks by maximizing mutual information between predictions on the original and masked graphs~\cite{ying2019gnnexplainer}.
(2) PGExplainer (\pgexp) trains an MLP as a mask generator on node embeddings, also optimizing mutual information~\cite{luo2020parameterized}.
(3) \cff learns feature and edge masks by optimizing a weighted combination of factual and counterfactual measures~\cite{tan2022learning}.
(4) \moe produces Pareto-optimal explanations balancing simulatability and counterfactual relevance~\cite{liu2021multi}; 
without constraint $k$, it may return up to $\sim$200 subgraphs.
(5) SAME develops structure-aware Shapley-based explanations via connected substructures~\cite{ye2023same}.

\begin{figure*}[tb!]
    \centering

    \begin{subfigure}[b]{0.24\linewidth}
        \includegraphics[width=\linewidth]{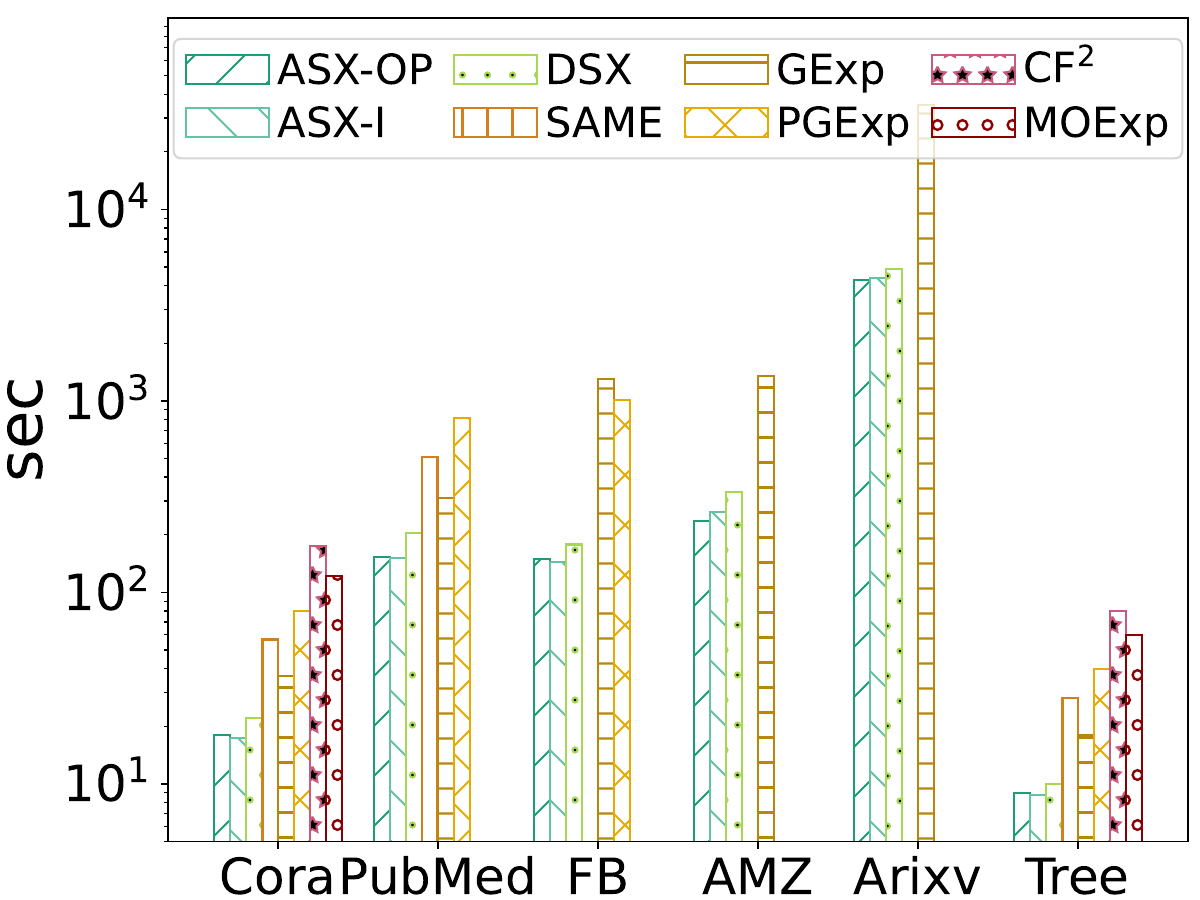}
        \caption{Efficiency}
        \label{fig:effi}
    \end{subfigure}
    \hfill
    \begin{subfigure}[b]{0.24\linewidth}
        \includegraphics[width=\linewidth]{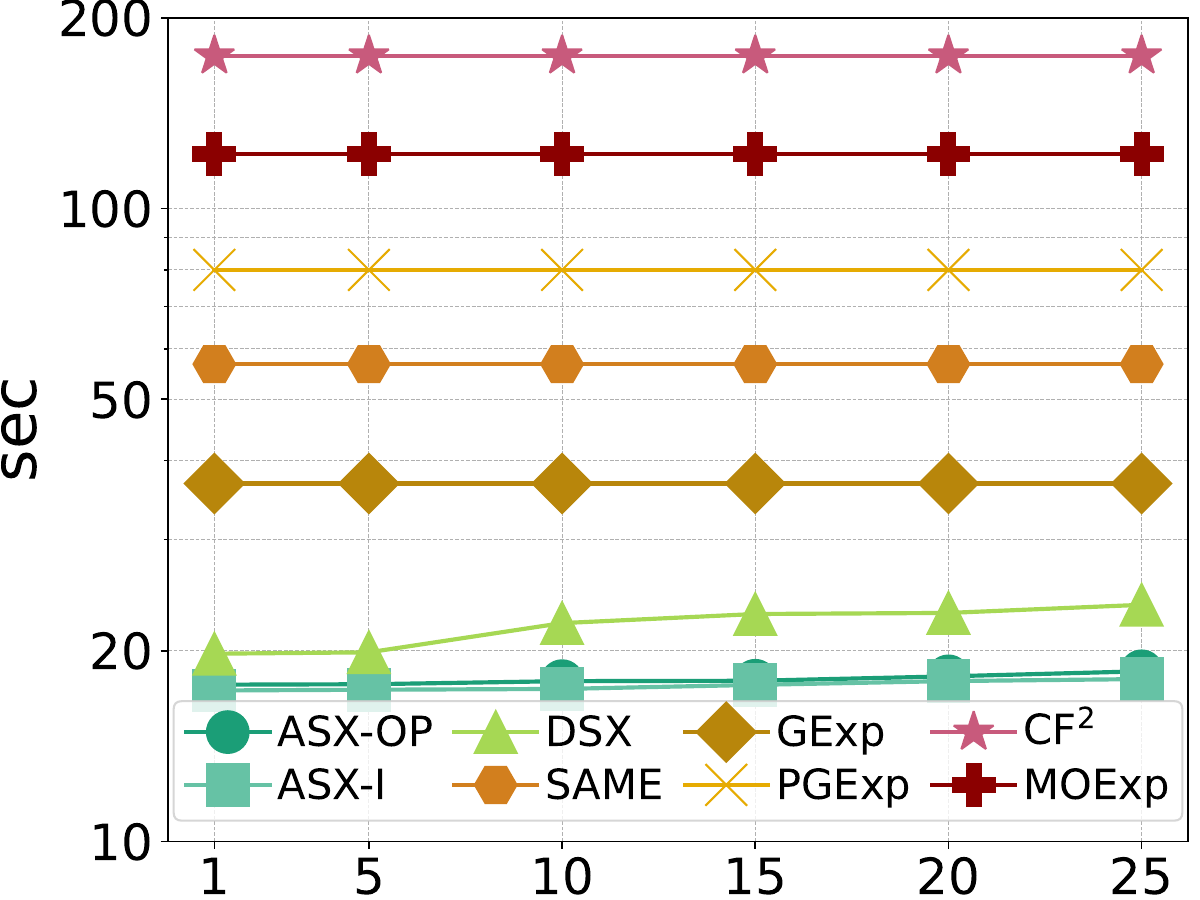}
        \caption{Scalability w.r.t. $k$ (\cora)}
        \label{fig:scale_k}
    \end{subfigure}
    \hfill
    \begin{subfigure}[b]{0.24\linewidth}
        \includegraphics[width=\linewidth]{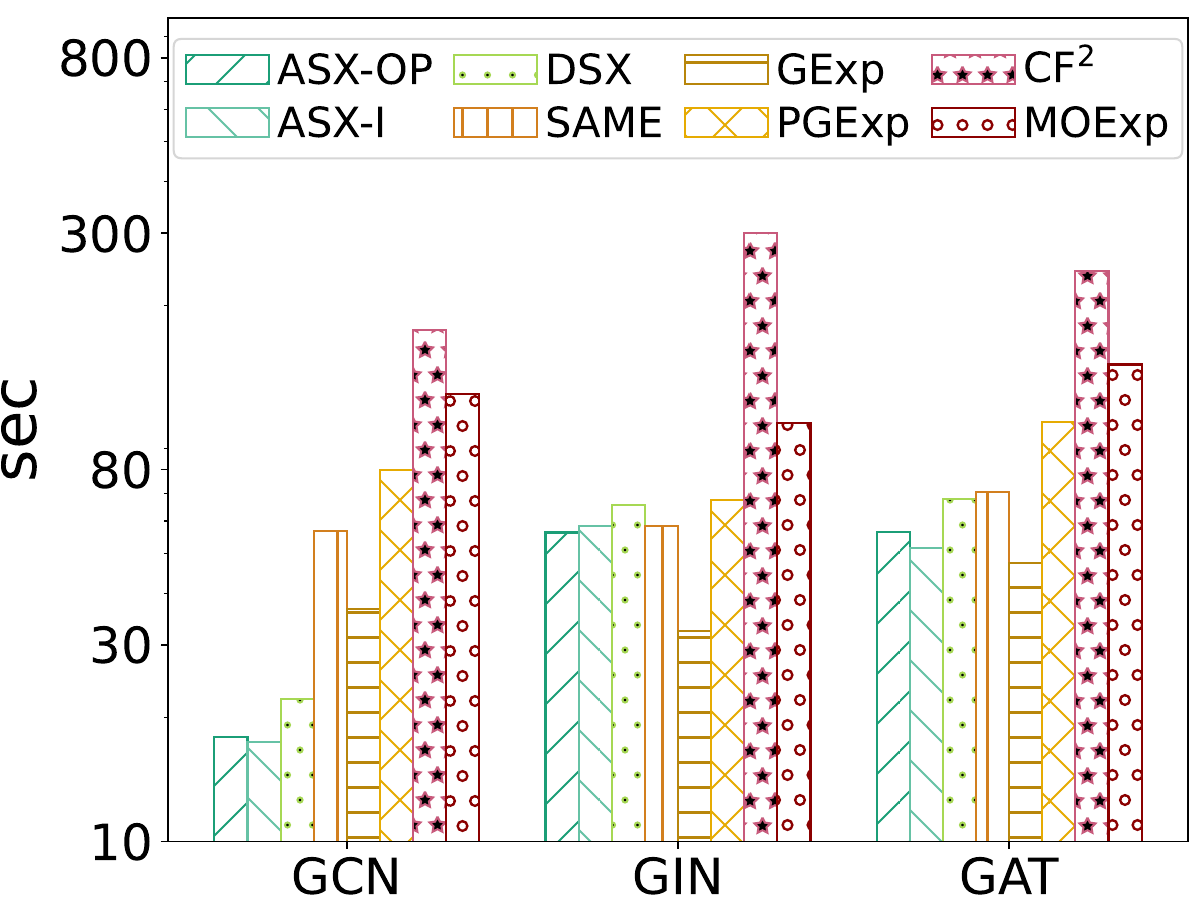}
        \caption{Scalability w.r.t. \gnns (\cora)}
        \label{fig:scale_gnn}
    \end{subfigure}
    \hfill
    \begin{subfigure}[b]{0.24\linewidth}
        \includegraphics[width=\linewidth]{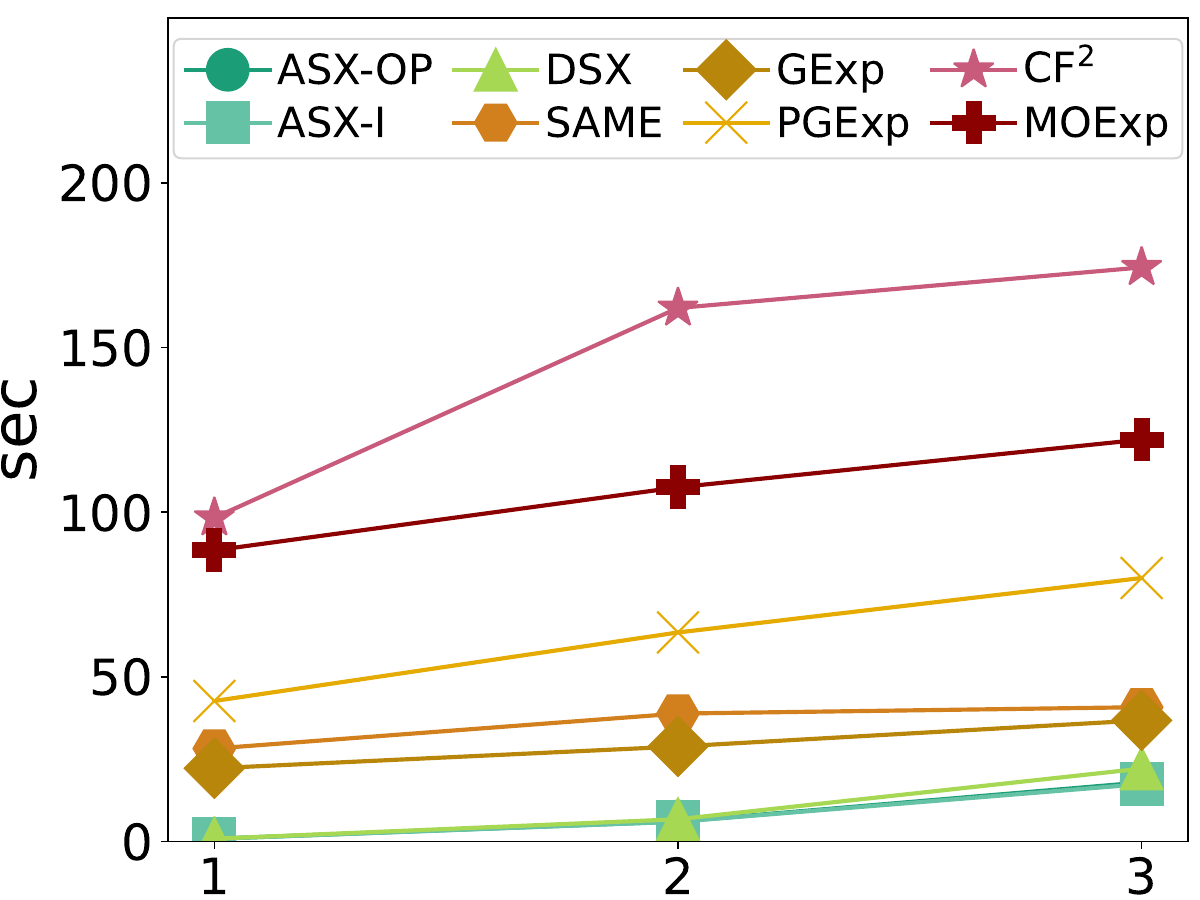}
        \caption{Scalability w.r.t. $L$ (\cora)}
        \label{fig:scale_l}
    \end{subfigure}
    
    \caption{Efficiency and scalability of our skyline explanation vs. SOTA explainers (\gnnexp, \pgexp, \cff, \moe, \kw{SAME}).}
    \label{fig:effi_scale}
    \vspace{-3ex}
\end{figure*}

\stitle{Evaluation Metrics.} 
For all datasets and \gnns, 
we select three common explainability 
measures ($\Phi$): $\fplus$, $\fminus$, 
and \conc. 
In addition, we adopt the explanation accuracy~\cite{ying2019gnnexplainer} (\acc) for the synthetic \treecycle dataset. \acc measures the proportion of human-crafted ground-truth nodes present in the explanatory subgraph. 
For computational simplicity, we transformed each minimization measure ($\conc$, $\fminus$) into its inverse counterpart, so that all measures are expressed as \emph{maximization} objectives.
As most \gnn explainers   
are not designed to generate explanations for 
multiple explainability measures, 
for a fair comparison, 
we employ three quality indicators (\textbf{QIs}) 
~\cite{li2019quality, qi2, qi3, qi4}. 
These quality indicators are widely-used to measure how good each result 
is in a multi-objective manner. 
Consider a set of $n$ \gnn 
explainers, where each explainer $A$ reports a 
set of explanatory subgraphs $\G_A$. 

\sstab
(1) \textbf{QI-1}: Integrated Preference Function (IPF) (Higher is better)~\cite{carlyle2003quantitative}. 
IPF score unifies and compares the quality of non-dominated set solutions with a weighted linear sum function. 
We define a normalized IPF of an explanation $\G_A$ from each \gnn explainer $A$ with a normalized 
single-objective score: 

\begin{small}
\begin{equation}
    \kw{nIPF}(\G_A) = \frac{1}{|\G_A|\cdot|\Phi|}\sum_{G\in\G_A}\sum_{\phi\in\Phi} \phi(G)
    \label{eq:wss}
\end{equation}
\end{small}

\sstab 
(2) \textbf{QI-2}: Inverted Generational Distance (IGD)
(Lower is better)
~\cite{coello2004study,li2019quality}, 
a most commonly used distance-based QI. It 
measures the distance from each solution 
to a reference set that contains 
top data points with  
theoretically achievable ``ideal'' values. We introduce IGD 
for explanations as follows. 
(a) We define a universal 
space $\G$ = $\bigcup_{i\in[1,n]}\G_i$ 
from all the participating \gnn explainers, 
and for each 
$\phi\in\Phi$, 
induces a reference set 
$\G_\phi^{k}\in\G$ with 
explanatory subgraphs 
having the top-$k$ values in $\phi$. 
(b) The normalized IGD of an 
explanation $\G_A$ from an explainer $A$ 
is defined as: 
\begin{small}
\begin{equation}
\kw{nIGD}(\G_A) = \frac{1}{k\cdot|\Phi|}\sum_{\phi\in\Phi}\sum_{G'\in\G_\phi^{k}}\min_{G\in\G_A} d(\phi(G), \phi(G'))     
\end{equation}
\end{small}
$d(\cdot)$ is Euclidean distance function following~\cite{li2019quality, qi2}.

\sstab 
(3) \textbf{QI-3}: Maximum Spread (MS) (Higher is better)~\cite{li2019quality}. \kw{MS} 
is a widely-adopted spread indicator that 
quantifies the range of the minimum and maximum values a solution can achieve in 
each objective. For a fair comparison, 
we introduce a normalized \kw{MS} score 
using reference sets in QI-2. 
For each measure $\phi\in\Phi$, 
and an explanation $\G_A$, 
its normalized MS score on $\phi$ is computed as: 
\begin{small}
\begin{equation}
\kw{nMS}(\G_A)^\phi = \frac{\phi(G^{A*}_\phi)}{\phi(G_\phi^*)}
\end{equation}
\end{small}
where $G_\phi^*$ is the explanatory subgraph with the best score on $\phi$ 
in universal set $\G$, and $G^{A*}_\phi$ 
is $\G_A$'s counterpart on $\phi$. 

\sstab 
(4) \textbf{Efficiency}: We report the total 
time of explainers. For learning-based methods, we include learning costs.

\subsection{Experimental Results}
\vspace{-1ex}

\stitle{Exp-1: Effectiveness}. We  
evaluate the overall performance of 
the \gnn explainers using QIs. 

\eetitle{QI-1: IPF Scores}. 
We report IPF scores (\textit{bigger} is better) for all \gnn explainers in Figure~\ref{fig:ipf_effect} with \gcns and $k$=$10$. Additional explainability results with varying $k$ are given in Appendix.
(1) In general, \apxsx outperforms a majority
of competitors 
in multi-objective explainability evaluation metrics. 
For example, on \cora, \apxsx outperforms \gnnexp, \pgexp, \cff, \moe, and \kw{SAME} 
in IPF scores by 1.34, 1.51, 2.81, 1.79, and 1.23 times, respectively. 
\divsx and \apxsxi achieve comparable performance with \apxsx. 
(2) 
We also observe that 
\gnnexp\ and \pgexp\ are sensitive as the datasets vary. 
For example, on \facebook, both show a significant change in IPF scores. 
In contrast, \apxsx, \apxsxi, and \divsx consistently 
achieve top IPF scores over all datasets. 
This verifies 
their robustness
in generating 
high-quality explanations over different 
data sources. 

\eetitle{QI-2: IGD Scores}. 
Figure~\ref{fig:igd_effect} reports the 
IGD scores (\textit{smaller} is better) of the explanations 
for \gcn-based classification. 
\apxsx, \apxsxi, and \divsx achieve the best in IGD scores among  
all \gnn explainers, for all datasets. 
This verifies that our explanation 
method is able to consistently select 
top explanations from a space of 
high-quality explanations that are separately 
optimized over different measures.
In particular, we find that 
\apxsx, \apxsxi, and \divsx are able to 
``recover'' the top-$k$ global 
optimal solutions for each 
individual explainability measures with 
high hit rates ($\approx$ 90\% of the cases). 

\eetitle{QI-3: nMS Scores}. 
Figure~\ref{fig:radar1} visualizes the 
normalized MS scores of the \gnn explainers, 
on \cora\ dataset, where $k$=$10$. 
(1) \apxsx reports a large range and contributes to 
explanations that are either close or the exact  
optimal explanation, for each of the three measures. 
\apxsxi and \divsx have comparable performance, yet with larger 
ranges due to its properly diversified solution. 
(2) \divsx, \apxsx, and \apxsxi contribute to the optimal 
$\fplus$, $\fminus$, and $\conc$, respectively. 
On the other hand, 
each individual explainer performs worse for 
all measures, with a large gap. For example, in \cora,
\moe only achieves up to 3\% of the best explanation (\divsx) over $\fminus$. 

\eetitle{IPF scores over subsets of $\Phi$}.
Figure~\ref{fig:radar2} presents a lattice graph where each node represents a subset of explainability measures ($\Phi$), and each edge corresponds to the removal of one measure $\phi$ from a top-down perspective. When $|\Phi|$=$1$, we observe that $\conc$ and $\fminus$ achieve the highest IPF scores, suggesting that they are the most discriminative measures in terms of dominance. In contrast, $\acc$ and $\fplus$ obtain relatively low IPF scores, indicating weaker dominance capability. For instance, the subset \{$\fplus,\acc$\} yields the lowest IPF score of $0.20$, whereas the subset \{$\fminus,\conc$\} achieves the highest score $0.94$, reflecting stronger dominance power compared to the other two measures.

\stitle{Exp-2: Efficiency}.
Using the same setting as in Figure~\ref{fig:overall_effect}, 
we report the time cost. 
Figure~\ref{fig:effi} exhibits the following. 

\sstab 
(1) \apxsx, \apxsxi, and \divsx outperform all 
(learning-based) \gnn explanations. 
\apxsx (resp. \divsx) on average outperforms \gnnexp, \pgexp, \cff, \moe, and \kw{SAME}
by 2.05, 4.46, 9.73, 6.81, and 3.17 (resp. 1.62, 3.61, 7.88, 5.51, and 2.57) times, respectively. 
 Moreover, \apxsx surpass 
\gnnexp, \pgexp, and \kw{SAME} over 
larger graphs. 
\cff and \moe fail to generate explanations 
due to high memory 
cost and long waiting time. 
Learning costs remain their major 
bottleneck. 

\sstab 
(2) \apxsx, \apxsxi, and \divsx are 
feasible in generating high-quality 
explanations for \gnn-based classification. 
For example, for \facebook\ with 22,470 nodes and 342,004 edges, 
it takes \apxsx around 
$150$ seconds to generate skyline explanations with 
guaranteed quality. 
This verifies the
effectiveness of its 
onion-peeling 
strategies. 

\sstab 
(3) 
\divsx does not incur significant overhead despite of diversified evaluation, because
the benefit of prioritization carries over to the diversified search. The incremental maintenance 
of $k$-explanations further reduces unnecessary 
verification.

\sstab 
(4) 
\apxsxi may outperform \apxsx 
for cases when 
the test nodes have ``skewed'' edge distribution in $L$-hop subgraphs, i.e., less direct neighbors but more ``remote'' neighbors, which favors the edge growth strategy of \apxsxi. 
\begin{figure*}[tb!]
    \centering
    \begin{subfigure}[b]{0.18\linewidth}
        \includegraphics[width=1.0\linewidth]{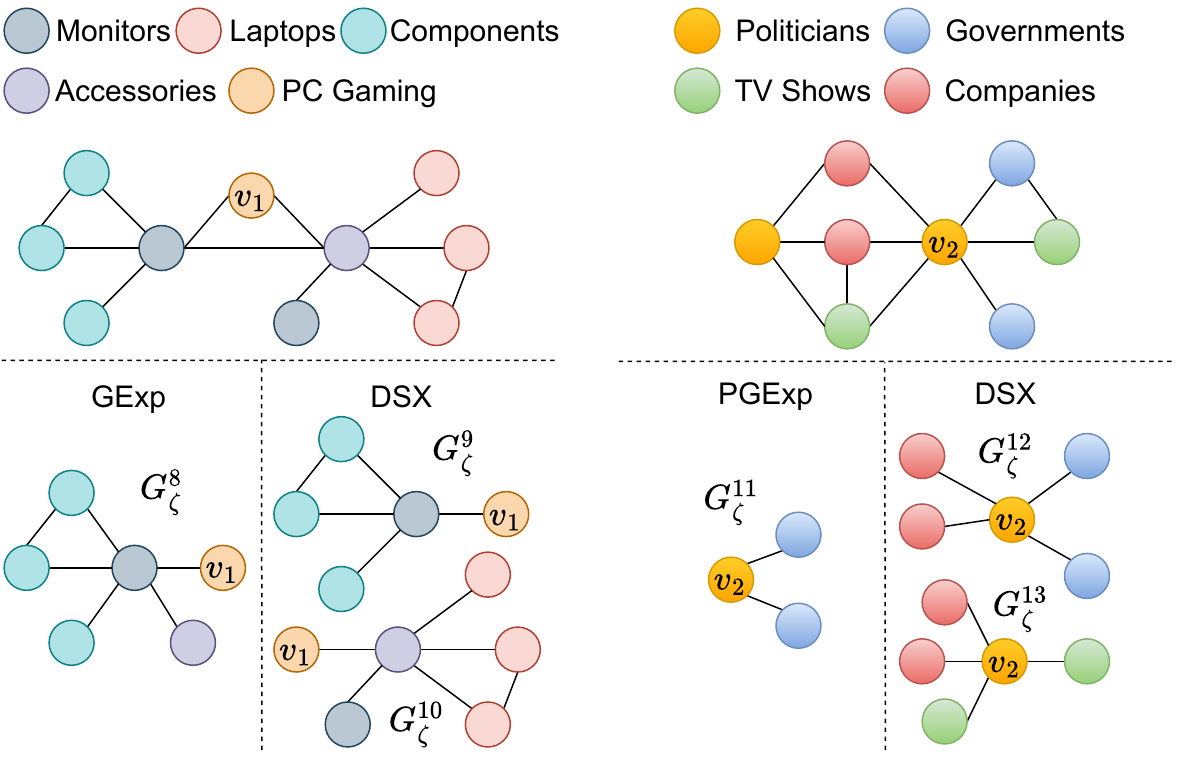}
        \caption{Comparison of different explanations on \amazoncomputer. }
        \label{fig:amazon_graph}
    \end{subfigure} 
    \hfill
    \begin{subfigure}[b]{0.26\linewidth}
        \includegraphics[width=1.0\linewidth]{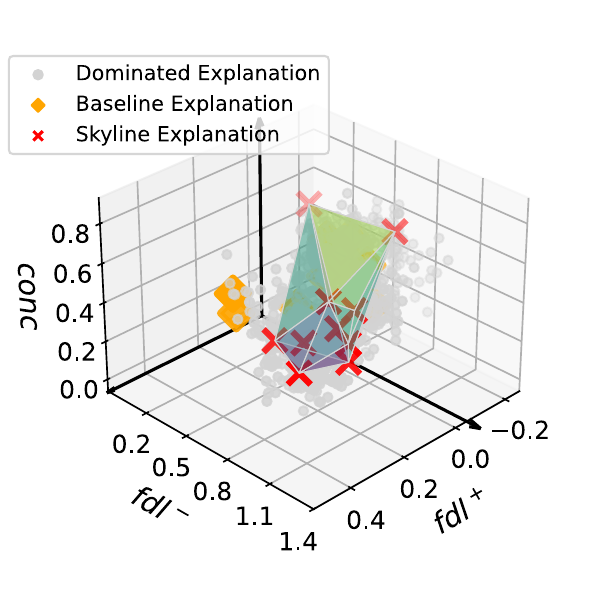}
        \caption{Dominance Coverage of Skyline Explanation v.s. Baseline Explanation (\amazoncomputer).}
        \label{fig:amazon_3d}
    \end{subfigure}   
    \hfill
    \begin{subfigure}[b]{0.18\linewidth}
        \includegraphics[width=\linewidth]{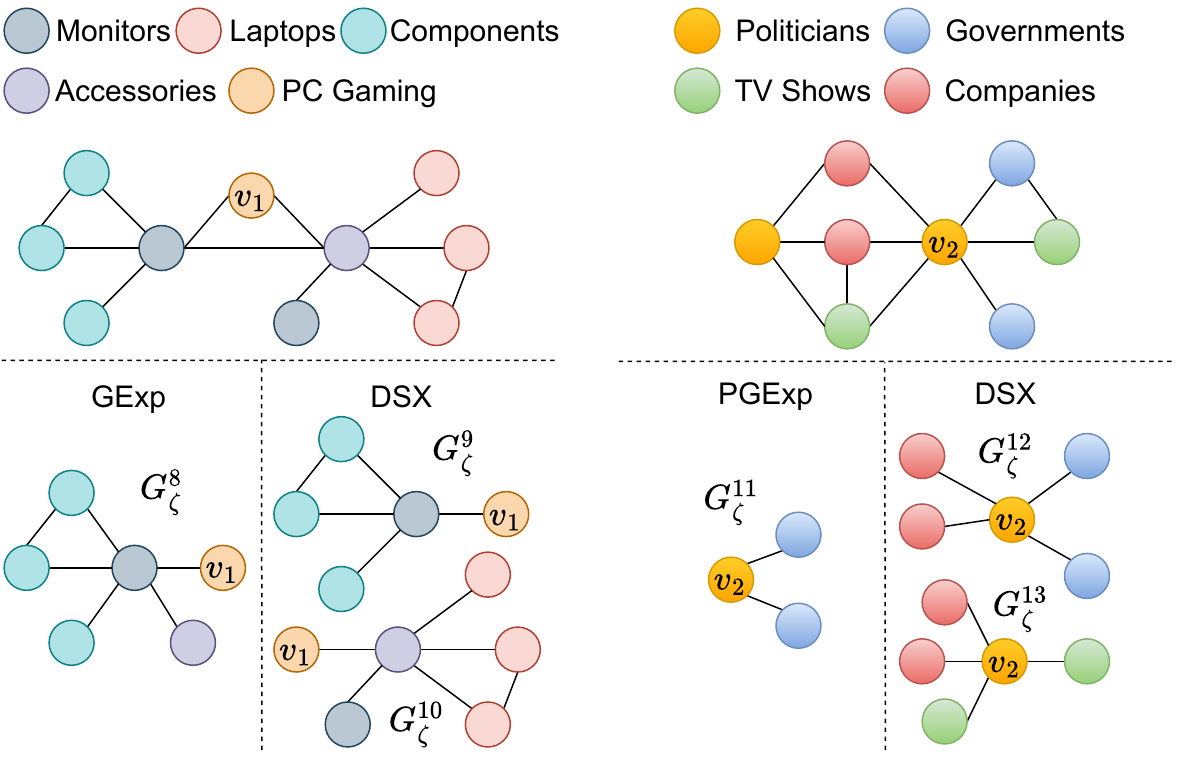}
        \caption{Comparison of different explanations on \facebook. }
        \label{fig:facebook_graph}
    \end{subfigure}
    \hfill
    \begin{subfigure}[b]{0.26\linewidth}
        \includegraphics[width=\linewidth]{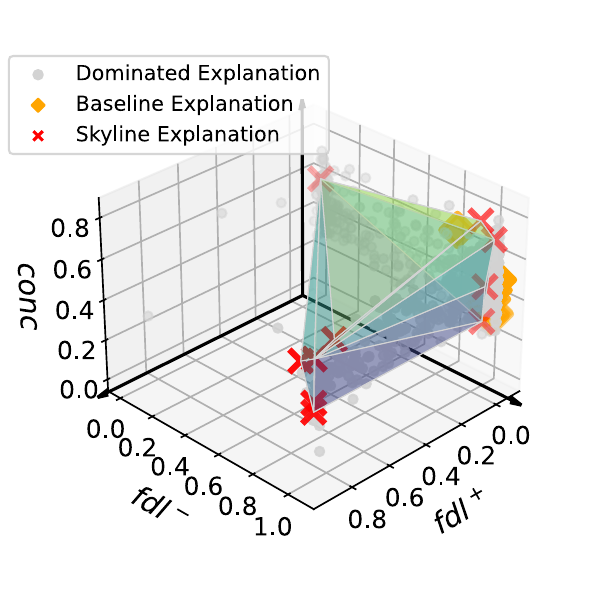}
        \caption{Dominance Coverage of Skyline Explanation v.s. Baseline Explanation (\facebook).}
        \label{fig:facebook_3d}
    \end{subfigure}
    \caption{Qualitative analysis of our skyline explanation and baseline explainers (\gnnexp, \pgexp, \cff, \moe, \kw{SAME}).}
    \vspace{-3ex}
    \label{fig:3d_sky}  
\end{figure*}

\stitle{Exp-3: Scalability.} 
We report the impact of critical factors (i.e., number of explanatory subgraphs $k$, \gnn classes, and number of \gnn layers $L$) on the scalability of 
skyline explanation generation, using   
\cora\ dataset. 
Additional results about effectiveness by varying factors are in Appendix.

\eetitle{Varying $k$}. 
Setting $\M$ as \gcn-based classifier with $3$ layers, 
we vary $k$ from 1 to 25. 
Since our competitors are not configurable w.r.t. $k$, we show the time costs of generating their solutions (that are independent of $k$), along with the time cost of our \apxsx, \apxsxi, and \divsx (which are dependent on $k$) in Figure~\ref{fig:scale_k}. 
Our methods take longer time to maintain skyline explanation with 
larger $k$, as more comparisons are required per newly generated 
candidate. 
\divsx is relatively more sensitive to $k$ due to additional computation of cosine
distances (\S\ref{sec-divsx}), yet remains 
faster 
than learning-based explainers. On average, our methods
take up to 23 seconds to maintain 
the explanations with $k$ 
varied to 25. 

\eetitle{Varying \gnn classes and $L$}. 
(1) Fixing $k$=$10$, 
we report the time cost of $3$-layer \gnn explainers 
for \gcn, \gat, and \gin 
over \cora.  As shown in Figure~\ref{fig:scale_gnn}, 
all our skyline methods take 
the least time to explain \gnn-based classification. 
This is consistent with our observation 
that the verification of \gnn is the most 
efficient among all classes, indicating an overall 
small verification cost. 
(2)
We fix $k$=$10$ and report the time cost of \gcn explainers, with the number of layers $L$ varied from 1 to 3 over \cora.  Figure~\ref{fig:scale_l} shows us that all our methods significantly outperform the competitors. The learning overheads of competitors remain their major bottleneck, while our algorithms, as post-hoc explainers without learning overhead, are more efficient. As expected,  all our methods take a longer time to generate 
explanations for larger $L$, as more subgraphs need to be verified from 
larger induced $L$-hop neighbors. 

\begin{figure}[tb!]
    \centering

    \begin{subfigure}[b]{0.48\linewidth}
        \includegraphics[width=\linewidth]{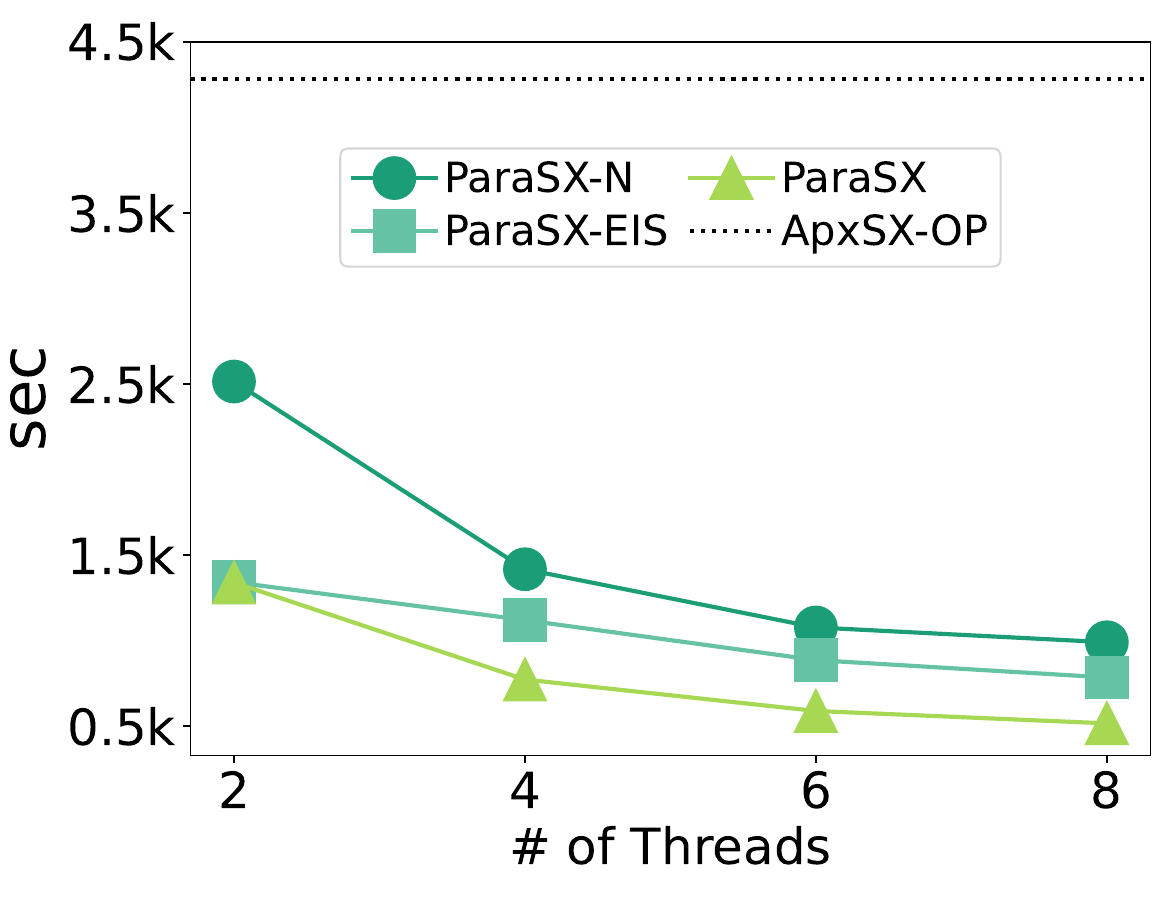}
        \caption{\arxiv}
        \label{fig:para_arxiv}
    \end{subfigure}    
    \hfill
    \begin{subfigure}[b]{0.48\linewidth}
        \includegraphics[width=\linewidth]{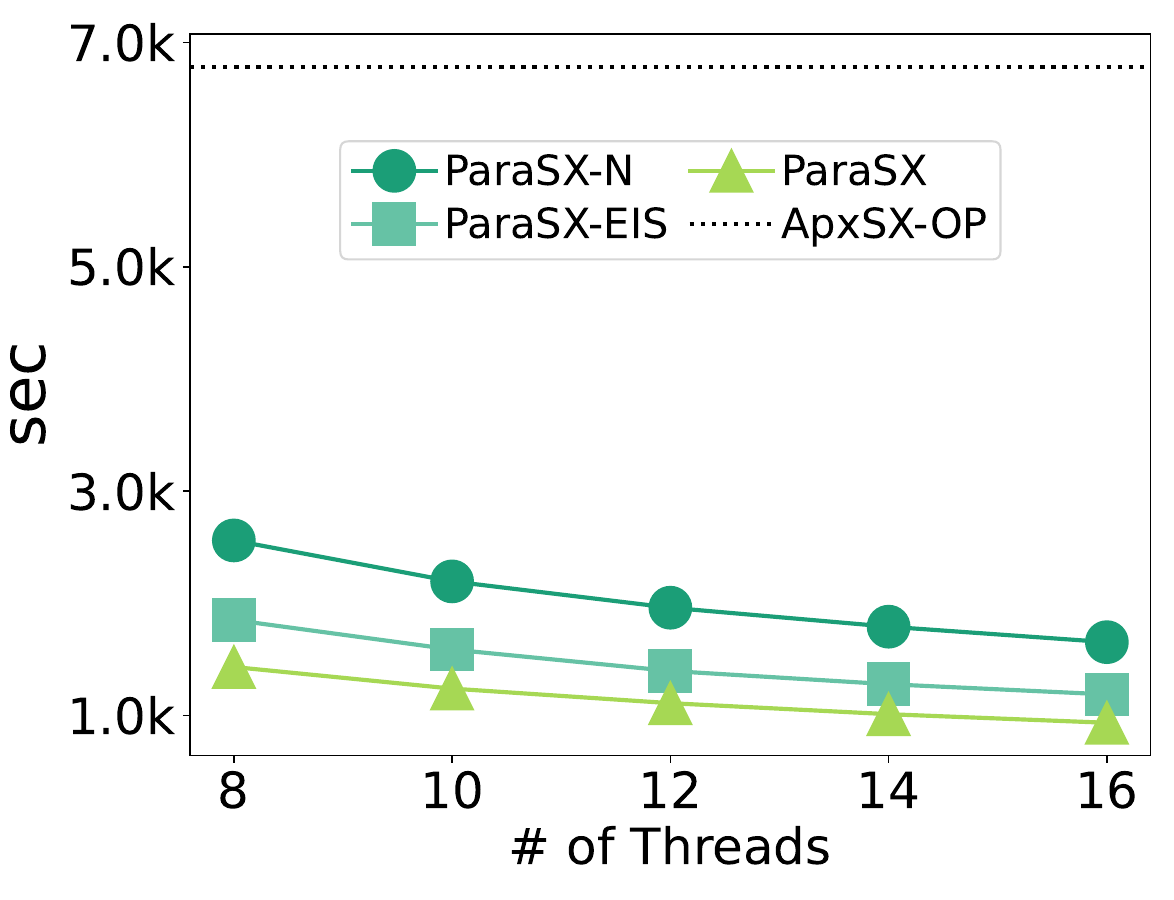}
        \caption{\bahouse}
        \label{fig:para_syn}
    \end{subfigure}
    \caption{Scaling to million-scale and billion-scale graphs.}
        \vspace{-3ex}
    \label{fig:para}
\end{figure}

\eetitle{Parallelization}. 
Setting $\M$ as \gcn-based classifier with $2$ layers, 
we vary the number of threads $m$ from $2$ to $8$ for \arxiv and $8$ to $16$ for \bahouse, respectively. 
We show the makespan of processing a query workload of $1000$ \sgxq for \arxiv, and $2000$ \sgxq for \bahouse, respectively. 
As show in Figure~\ref{fig:para} (with the dashed horizontal line showing the cost of single-thread processing \apxsx), 
our skyline explanation scheme can be effectively parallelized 
to million-scale and billion-scale graphs, and scales well 
as more threads are used. \parasx improves the 
efficiency of \parasxn and \parasxeis by 
$1.8$ and $1.4$ times on average, due to load balancing 
and scheduling strategies. 

\stitle{Exp-4: Case Analysis}. 
We next showcase   
qualitative analyses of our explanation methods, using real-world examples from two datasets: \amazoncomputer\ and \facebook. 

\eetitle{Diversified Explanation}. 
A user is interested in finding ``Why'' 
a product $v_1$ is labeled ``PC Gaming'' by a 
\gcn. \gnn explainers that 
optimize a single measure (\eg \, \gnnexp) 
 return explanations (see $G_{\zeta}^8$ in Figure~\ref{fig:amazon_graph}) that 
simply reveals a fact that 
$v_1$ is co-purchased with 
``Components'', a ``one-sided'' 
interpretation. 
In contrast, \divsx 
identifies an explanation that reveals 
 more comprehensive interpretation 
with two explanatory subgraphs (both factual) 
$\{G_{\zeta}^9, G_{\zeta}^{10}\}$ (Figure~\ref{fig:amazon_graph}), 
which reveal two co-purchasing 
patterns bridging $v_1$ with not only 
``Components'' for ``Monitors'' (by $G_{\zeta}^9$), 
but also 
to ``Accessories'' of ``Laptops''.
Indeed, 
the former indicates gamers 
who 
prefer gaming PC with 
high-refresh rate monitors; 
and the latter indicates that $v_1$ is a 
gaming laptop in need of frequent maintenance 
with laptop accessories. This showcase
that \divsx provides more comprehensive explanation.

In Figure~\ref{fig:amazon_3d}, we visualize the distribution of explanatory subgraphs for $v_1$ in Figure~\ref{fig:amazon_graph}. Each point represents a subgraph with 3D coordinates given by $\fplus$, $\fminus$, and $\conc$ scores. The gray dots denote the verified explanatory subgraph in the interpretable space, while baseline explainers (\gnnexp, \pgexp, \cff, \moe, \kw{SAME}) are shown as diamonds. Our skyline explanations appear as red crosses, with their convex hull illustrating dominance coverage. We observe that skyline explanations span most of the interpretable space, offering comprehensive and diverse explanations, whereas baseline explanations cluster in smaller regions, missing diversity~\cite{das1997closer}.

\eetitle{Skyline vs. 
Multi-objective Explanation (Linear Combination).}
Our second case compares \divsx and \pgexp\ (shown in Figure~\ref{fig:facebook_graph}). The latter generates explanations by optimizing a single objective that combines two explanation measures (factuality and conciseness). 
We observe that \pgexp\ generates small factual subgraphs, such as $G_{\zeta}^{11}$ for a test node $v_2$, yet relatively less informative. \divsx generates a skyline explanation $\{G_{\zeta}^{12}, G_{\zeta}^{13}\}$ that is able to interpret 
$v_2$ as ``Politicians'' not only due to its connection with ``Governments” 
entities but also highly related to ``TV shows'' and ``Companies''. 
Meanwhile, Figure~\ref{fig:facebook_3d} shows that our skyline explanations span most of the interpretable space, providing diverse dominance coverage, whereas baseline explainers cluster their solutions within limited regions.

\eat{
Meanwhile, Figure~\ref{fig:facebook_3d} verifies that our skyline explanation covers most of the interpretable space with diverse and comprehensive explanations, while the baseline explainers cluster their solutions in smaller regions. 
}



\vspace{-2ex}
\section{Conclusion}
\vspace{-2ex}
We proposed a class of skyline explanations that optimize multiple explainability measures for interpreting GNN predictions. We showed the hardness of explanatory query processing problem. To address this, we developed sequential, parallel, and diversified algorithms with provable Pareto-optimality guarantees. Extensive experiments demonstrate that our methods are practical for large-scale graphs and GNN-based classification, producing more comprehensive explanations than state-of-the-art GNN explainers.

\stitle{Acknowledgment}.
Dazhuo Qiu and Arijit Khan acknowledge support from the Novo Nordisk Foundation grant NNF22OC0072415. 
Haolai Che and Yinghui Wu acknowledge support from the NSF OAC-2104007. 

\bibliographystyle{IEEEtran}
\bibliography{ref}

\begin{figure*}[ht]
    \centering
    
    \begin{subfigure}[b]{0.24\linewidth}
        \includegraphics[width=\linewidth]{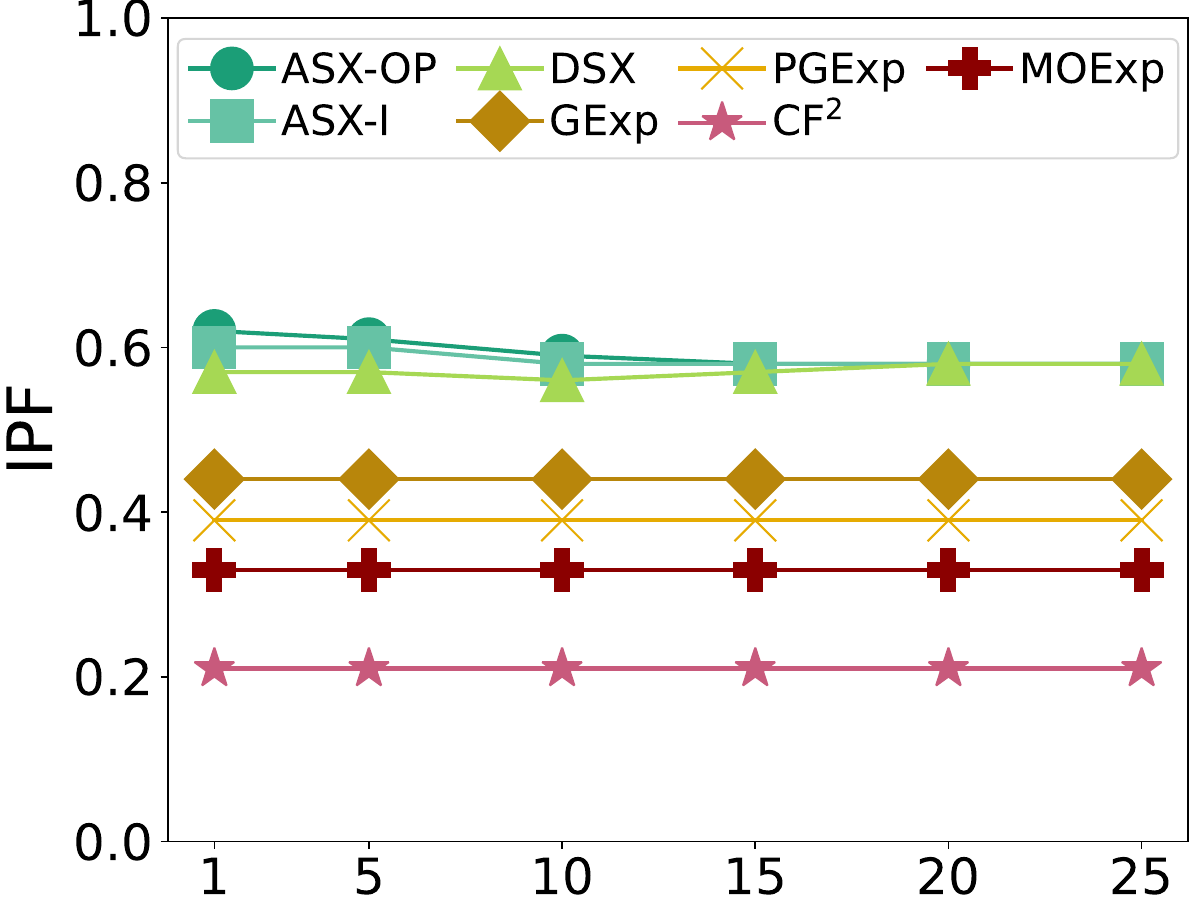}
        \caption{IPF varying $k$}
        \label{fig:ipf_k}
    \end{subfigure}    
    \hfill
    \begin{subfigure}[b]{0.24\linewidth}
        \includegraphics[width=\linewidth]{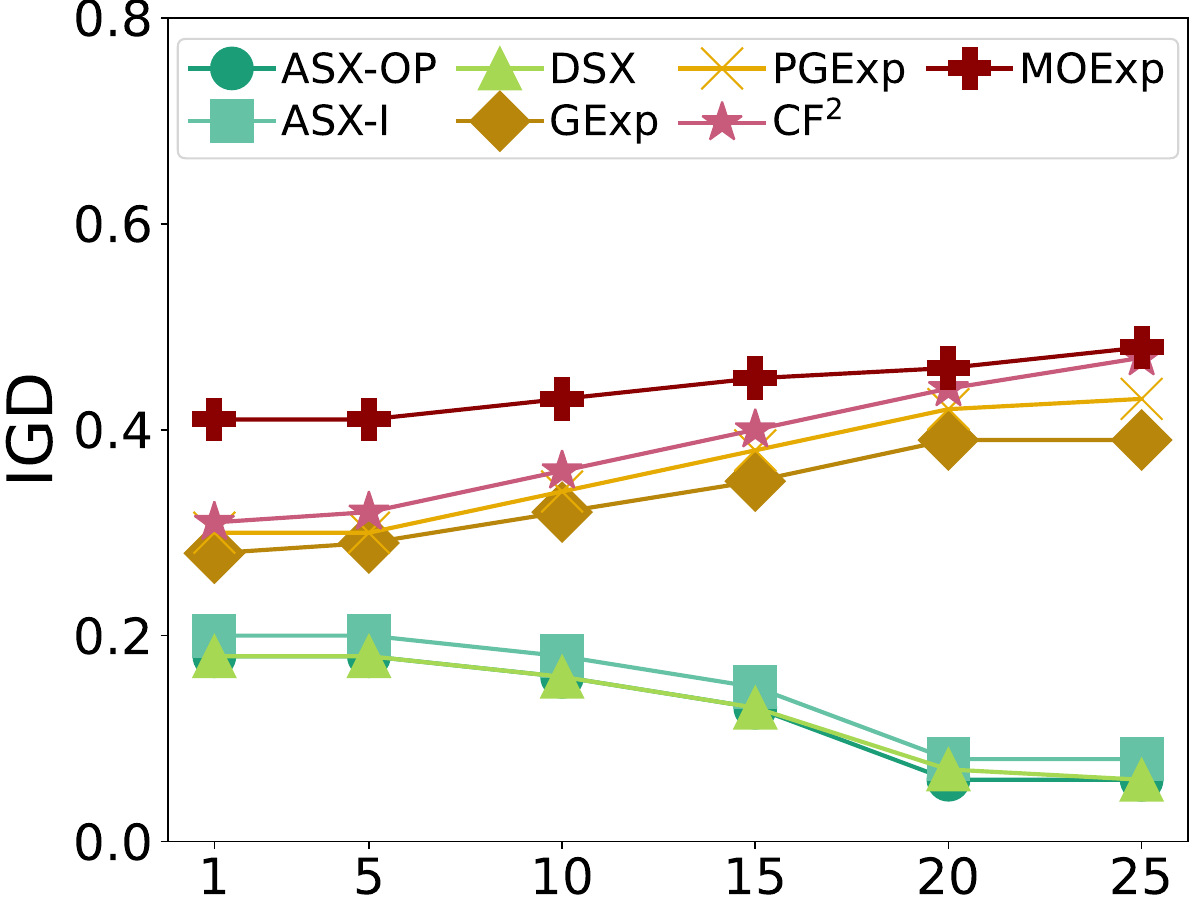}
        \caption{IGD varying $k$}
        \label{fig:igd_k}
    \end{subfigure}
    \hfill
    \begin{subfigure}[b]{0.24\linewidth}
        \includegraphics[width=\linewidth]{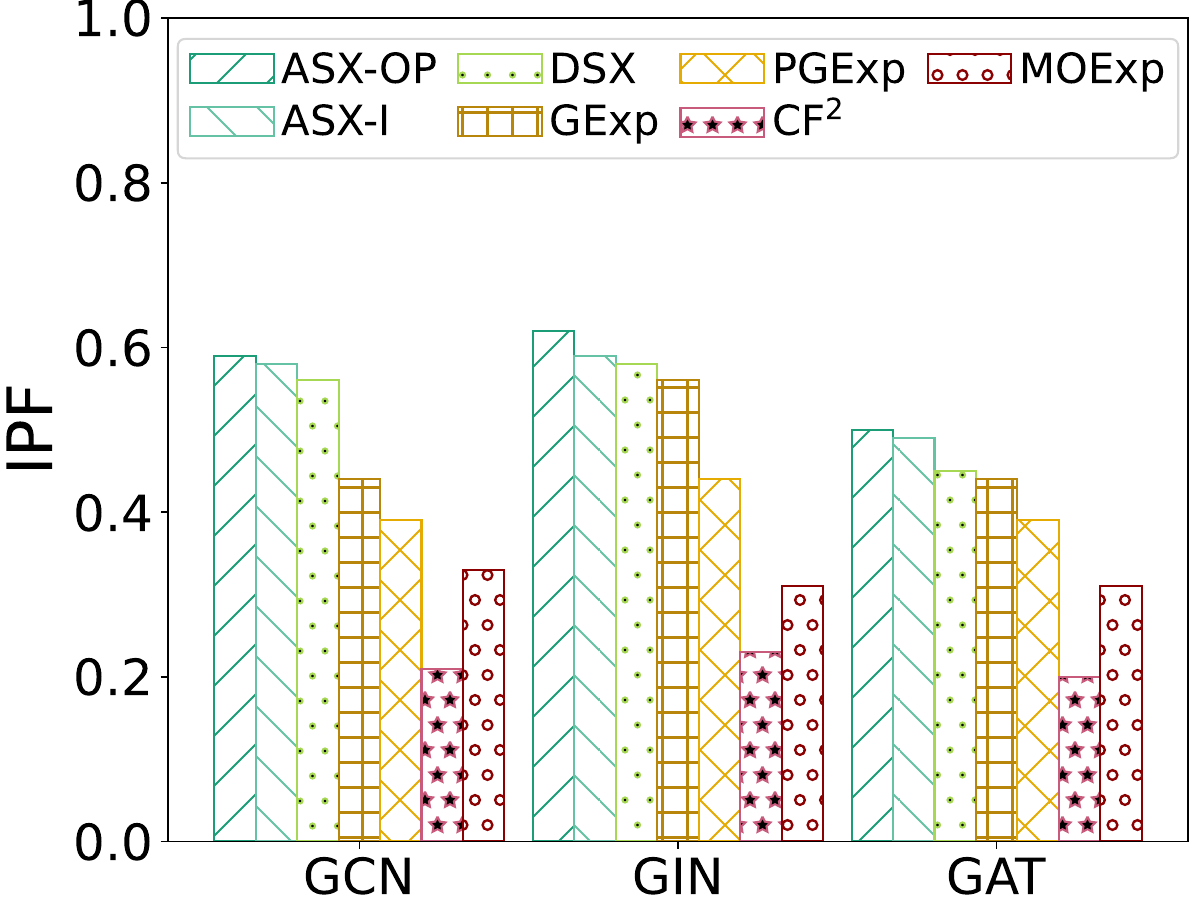}
        \caption{IPF varying \gnns}
        \label{fig:ipf_gnn}
    \end{subfigure}
    \hfill
    \begin{subfigure}[b]{0.24\linewidth}
        \includegraphics[width=\linewidth]{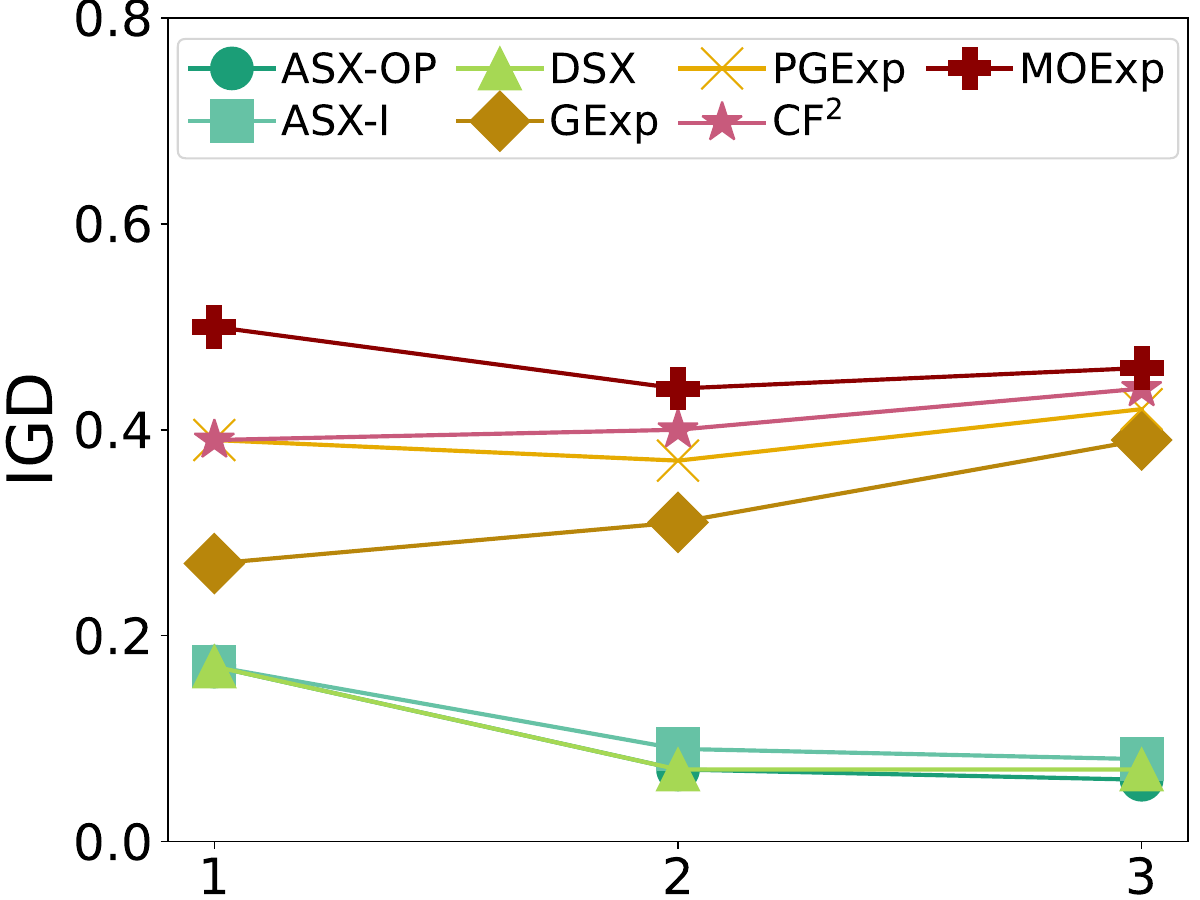}
        \caption{IGD varying $L$}
        \label{fig:ipf_L}
    \end{subfigure}

    \caption{Impact of factors on the effectiveness of explanations 
    }
    \label{fig:factor_effect}
\end{figure*}

\section*{Appendix}
\label{sec-appendix}

\subsection{Details of Algorithms and Procedures}

\stitle{Procedure \updatesx: Update Dominance Relation.}
We introduce the details of 
Procedure \updatesx; specifically, 
we describe how the (1+$\epsilon$) dominance relations 
of exploratory subgraphs are dynamically 
maintained. 
With the stream of upcoming states, i.e., explanatory subgraphs, we maintain a lattice structure for the dominance relationships. 
Specifically, we first compare each newly arrived state $s$ with the current skylines, 
and incrementally update the dominance lattice by conducting one of the following steps: 
\tbi 
\item If certain skylines (1+$\epsilon$)-dominate it, it inserts a directed edge from each dominating skylines to the new state; the checking of (1+$\epsilon$)-dominance takes constant time for small measure space involving several explainability measures; 
\item If the new state dominates certain skylines, we connect directed edges from the new state to each dominated skylines;
Meanwhile, we reconnect the dominant states of these skylines to the new state;   
\item If the new state is not dominated by any skyline, then we obtain its $\dset(s)$ using the bitvector $B(s)$. 
\ei 

Next, we identify the skyline $\sover$ with the smallest $\dset(\sover)$, replace $\sover$ with $s$, only when such replacement makes the $\dset$ increased by a factor of $\frac{1}{k}$. 
If swapping is conducted, the new skyline, i.e. $s$, connects directed edges to the dominating states based on bitvector $B(s)$, 
and we remove the swapped skyline and its corresponding edges. 
The above swap strategy ensures a $\frac{1}{4}$-approximation ratio~\cite{onlinemkc}. 

\stitle{Algorithm~\apxsxi: Alternative ``Edge Growing'' Strategy.}
As the end user may want early termination and obtain a compact, 
smaller-sized explanations, we also outline a variant of \apxsx, denoted as~\apxsxi. It follows 
the same \kw{Verifier} and \kw{Updater} 
procedures, yet uses a different 
\kw{Generator} procedure that 
starts with a single node $v$ 
and inserts edges to grow candidate, 
level by level, up to its $L$-hop 
neighbor subgraph. 
\apxsxi remains to be an 
$(\frac{1}{4},\epsilon)$-approximation for $\eval(\sgxq^k)$, 
and does not incur additional 
time cost compared with \apxsx. 

Edge growing strategy differs from the “onion peeling” strategy with the test node as the initial state and the $L$-hop neighbor subgraph as the final state. 
\apxsxi\ adopts Breadth-First Search to explore explanatory subgraphs. 
Specifically, starting from the test node, we first connect the test node with each of its neighbors, respectively. 
Then we verify these candidate subgraphs and determine the ($1+\epsilon$)-dominance relationships. 
Following the same candidate prioritization as \apxsx, we select the neighboring node with the biggest weight. 
Then, continuing with this newly chosen node, we explore its neighbors, except the ones that we previously visited. 
We expand the subgraph with the neighbors, respectively, and conduct candidate prioritization again.
We iteratively continue this process until we explore the explanatory subgraph as the $L$- neighbor subgraph. 
The search space and time complexity of \apxsxi\ remains the same as \apxsx, since the interpretation domain $\zeta'$ and the update strategy are the same.

\begin{algorithm}[tb!]
\renewcommand{\algorithmicrequire}{\textbf{Input:}}
\renewcommand{\algorithmicensure}{\textbf{Output:}}
\caption{\divsx\ Algorithm}
    \begin{algorithmic}[1]
        \Require 
        a query $\sgxq^k$ = $(G,\M, v_t,\Phi)$; 
        a constant $\epsilon\in[0,1]$; 
        \Ensure 
        a $(\zeta',\epsilon)$-explanation $\G_\epsilon$. 
         \State set $\G_\epsilon$:=$\emptyset$; 
         \State identify edges for each hop: $\eset$:=$\{E_L, E_{L-1},\ldots,E_1\}$; 
        \For{$l$ = $L$ to $1$} 
            \State initializes state $s_{0}$:=$G^l(v_t)$ 
            \While{$E_l\neq \emptyset$}
                \For{$e\in E_l$} 
                    \State spawns a state $s$ with candidate $G_s$:=$G^l\backslash \{e\}$; 
                    \State update $\zeta'$ with state $s$ and new transaction $t$;  
                    \If{$\veriF(s)$=False \& $\veriCF(s)$=False} 
                        \State continue;  
                    \EndIf 
                    \State 
                    $\G_\epsilon$:=\updatedivsx$(s, G_s,\zeta', \G_\epsilon)$; 
                    $E_l$=$E_l\backslash\{e\}$; 
                \EndFor
            \EndWhile 
        \EndFor\\
        \Return $\G_\epsilon$.    
    \end{algorithmic}
  \label{alg:divsx}

\vspace{1.5ex}

\textbf{Procedure} \updatedivsx
\begin{algorithmic}[1]
        \Require 
        a state $s$, candidate $G_s$, state graph $\zeta'$; explanation $\G_\epsilon$; 
        \Ensure Updated 
        $(\zeta', \epsilon)$-explanation $\G_\epsilon$;  
        \State 
        initializes state $s$ with structure $\Phi(G_s)$:=$\emptyset$;  
        \State evaluates $\Phi(G_s)$; \label{alg:div_phi}
        \State incrementally determines $(1+\epsilon)$-dominance 
        of $G_s$; \label{alg:div_dom}
        \If{$\{G_s\}$ is a new skyline explanation} 
        
        \State $\Delta(s|\G_\epsilon)$ := $\divs($$\G_\epsilon$$\cup$$\{G_s\}$$)- \divs($$\G_\epsilon$$)$\label{alg:div_margin}
            \If{$|\G_\epsilon|$$<$$k$ \textbf{and} $\Delta(s|\G_\epsilon)$$\geq$$\frac{(1+\epsilon)/2-\divs(\G_\epsilon)}{k-|\G_\epsilon|}$} \label{alg:div_thres}
                \State 
                $\G_\epsilon$ := $\G_\epsilon$$\cup$$\{G_s\}$; 
                \label{alg:div_add}
            \EndIf 
        \EndIf \label{cd-div-domd2}\\
        \Return $\G_\epsilon$.
    \end{algorithmic}
  \label{alg:updatedivsx}

\end{algorithm}

\stitle{Diversification Algorithm.}
The detailed pseudo-codes of diversification algorithm \divsx\ are shown as Algorithm~\ref{alg:divsx} and Procedure~\ref{alg:updatedivsx}. Please find the descriptions of \divsx\ in \S~\ref{sec-divsx}.

\eetitle{Approximability and Time cost of \divsx}. 
We can verify that $\ncs(\cdot)$ and $\cd(\cdot)$ are submodular functions. 
\eat{For $\ncs(\cdot)$, since we only modify the presence of edges and keep the nodes unchanged, when a new \warn{skyline} is included in the $k$-skyline explanation, the number of covered nodes will increase or remain the same. For $\cd(\cdot)$, when a new \warn{skyline} is included, the number of pair-wise cosine distance computations will increase by $|\G_\epsilon|$, therefore the overall $\cd(\cdot)$ will increase or remain the same.}
Consider procedure~\updatedivsx 
upon the arrival, at any time, 
of a new verified candidate $G_s$. 
Given that $|\G_\epsilon| \leq k$ 
is a hard constraint, 
we reduce the 
diversified evaluation of $\sgxq^k$ 
to an instance of the Streaming Submodular Maximization problem~\cite{ssm}. 
The problem maintains  
a size-$k$ set that optimizes a submodular function
over a stream of data objects. 
The diversification function $\divs$ is submodular, as it combines two submodular components: $\ncs$ and the summation over $\cd$. Specifically, $\ncs$, representing node coverage, is a submodular function bounded within $[0, N]$, with each incremental gain diminishing as more nodes are covered. In parallel, $\cd$ denotes the cosine distance between two embeddings, which is bounded in $[0, 1]$. Since pairwise cosine distance is a non-negative and bounded measure, the summation over $\cd$ across a set remains submodular. Consequently, their combination in $\divs$ preserves submodularity.
\divsx 
adopts a greedy 
increment policy by 
including a candidate in $\G_\epsilon$ with the new 
candidate $G_s$ only when 
this leads to a marginal gain greater than $\frac{(1+\epsilon)/2-f(S)}{k-|S|}$, where $S$ is the current set and $f(\cdot)$ is a submodular function. This is consistent with an increment policy that ensures a ($\frac{1}{2}-\epsilon$)-approximation in~\cite{ssm}. 
Since \divsx 
follows the same process as \apxsx but only differs in 
replacement policy with same time cost,
the cost of \divsx is also $O(|\zeta'|(\log\frac{r_\Phi}{\epsilon})^{|\Phi|}+|\zeta'|L|G^L(v_t)|)$.

\begin{algorithm}[tb!]
\renewcommand{\algorithmicrequire}{\textbf{Input:}}
\renewcommand{\algorithmicensure}{\textbf{Output:}}
\caption{\parasx\ Algorithm}
    \begin{algorithmic}[1]
        \Require 
        a set of queries $Q$ = $\{\sgxq^k_1, \sgxq^k_2, \cdots, \sgxq^k_n\}$; 
        the number of threads $m$;
        \Ensure 
        a set of $(\zeta',\epsilon)$-explanations $\gq$ for $Q$. 

        \State $\gq$ := $\emptyset$; $V_T$ := $\bigcup_{i=1}^{n} \sgxq^k_i.{v_t}$; node partition $P$ := $\emptyset$; 
        \State extract $L$-hop neighbors of $V_T$: $G_{V_T}$; globally share $G_{V_T}$, $\M$, $\Phi$; \label{pseudo-pre}

        \State $P$ := $\cp(V_T, G_{V_T})$;\label{pseudo-cp}

        \For{$i$ = $1$ to $m$}\label{pseudo-order1}
            \State order $p_i$ by computing node influential scores. \label{pseudo-order2}
            \State initialize edge information sharing table $eist^i$ := $\emptyset$; \label{pseudo-eist1}
            \For{$v_t \in p_i$}
                \State $\gq^i[v_t]$ := \apxsx($v_t$) \label{alg:apxsxop}
                \State update $eist^i$ with newly visited edges;
            \EndFor
        \EndFor\label{pseudo-eist2}
        \State $\gq$ := $\bigcup_{i=1}^{m} \gq^i$;\label{pseudo-final1}\\
        \Return $\gq$.\label{pseudo-final2}
    \end{algorithmic}
  \label{alg:para_sx}

\vspace{1ex}

\textbf{Procedure} \cp
\begin{algorithmic}[1]
    \Require 
    a set of test nodes $V_T$; 
    the subgraph of $V_T$: $G_{V_T}$;
    \Ensure 
    a partition $P$ of $V_T$. 

    \State generate MinHash signatures for $V_T$;
    \State indexing signatures by Locality-Sensitive Hashing;
    \State partition $V_T$ into $m$ subsets based on LSH indexing: $P$ := $\{p_1, p_2, \cdots, p_m\}$;
    \State Balance the sizes of the partitions;\\
    \Return $P$.
\end{algorithmic}
\label{alg:cp}
\end{algorithm}

\stitle{Parallel Algorithms}. 
We first present the details of the baseline parallel algorithms (\parasxn and \parasxeis). They 
differ in modeling and minimizing the makespan of skyline explanatory query processing, i.e., the total elapsed (wall-clock) time from the start of the first $\sgxq^k$ to the completion of the last $\sgxq^k$ in a batch. 
Our parallel algorithms aim to tackle the common makespan minimization problem. 
Given a batch of $\sgxq^k$ queries $Q$ = $\{\sgxq^k_1, \sgxq^k_2, \cdots, \sgxq^k_n\}$ and a set of threads running in parallel: $T$ = $\{t_1, t_2, \cdots, t_m\}$, the objective is to assign each thread $t_i$ a subset of query workload $Q_i\subseteq Q$, such that the maximum time of any thread takes to complete its subset of queries (or its makespan) is minimized. More formally, it computes an assignment $\Psi$ : $Q$ $\rightarrow$ $T$ to minimize
\[
\max_{t\in T} \sum_{\sgxq^k\in\Psi(t)} c(\sgxq^k)
\]
where $c(\cdot)$ is the (estimated) execution time of a given $\sgxq^k$. 

\eetitle{A Naïve solution.} 
An intuitive approach to minimize the $\sgxq^k$ makespan is randomly splitting $Q$ into $\frac{n}{m}$ subsets and assign each of them to one thread. Within each thread, we can select one skyline generation algorithm of interest: \apxsx, \apxsxi, or \divsx, then invoke the algorithm to iteratively generate the skyline explanation for each query. We refer to this baseline approach as \parasxn.

\eetitle{An improved approach}. We next introduce an improved approach, 
denoted as \parasxeis. The partition strategy of the queries is the same as \parasxn--the batch of queries is randomly split and assigned. It differs from \parasxn 
in using a shared edge table to avoid unnecessary computation 
for common edges in L-hop neighbors of the test nodes; and 
adopts a node prioritization strategy to dynamically process influencing nodes.

\stab 
(1) Edge Information Sharing. Considering a subset of skyline explanatory queries assigned to a thread, the computation cost boils down to the computation of the corresponding subset of test nodes. We observe that test nodes often share common neighbors. As noted in Section~\ref{Optimization}, previous optimization techniques compute marginal gains per edge for each individual test node, resulting in redundant computations across overlapping neighborhoods. To mitigate this, we propose an approach that maintains a per-thread lookup set to track visited edges across all test nodes assigned to the same thread.

\stab 
(2) Influential Node Prioritization. 
Firstly, since edge information is computed with respect to individual test nodes, any shared edge is attributed to the first node that processes it. To ensure this shared information is representative of other test nodes within the same thread, it is important to prioritize nodes that are more influential or centrally located among the assigned subset. To this end, we propose an influential node prioritization strategy that ranks test nodes within each thread, ensuring that nodes with higher centrality are processed earlier.
Each thread begins by applying the $k$-core decomposition algorithm~\cite{core_decomp} to assign a core number to each node, reflecting the highest $j$ such that the node belongs to the $j$-core. Test nodes are then ranked in descending order of core number, prioritizing more “central” nodes in the thread. This ordering ensures that edge information from structurally important nodes is shared early, maximizing reuse.


Bringing these strategies into the algorithm at runtime, we first perform node prioritization (as described in \S~\ref{edge-prior}), i.e., the most influential node in the thread. For each edge in its neighborhood, we compute the marginal gain, determine the deletion direction using the ``onion-peeling'' strategy, and store the resulting edge scores in the lookup set. For subsequent test nodes, any edge already recorded in the lookup set is skipped, and its previously computed marginal gain is directly reused. Only unseen edges are processed and added to the set, thereby reducing redundant computation. 

\stitle{Algorithm \parasx.}
Recall our ultimately proposed parallel algorithm \parasx in Section~\ref{sec-parallel}. The key distinction between \parasx and its two baseline variants (\parasxn and \parasxeis) lies in its novel partitioning strategy, which enables more effective node clustering and promotes greater overlap in shared neighborhoods among nodes within each thread.

As illustrated in Algorithm~\ref{alg:para_sx}, we begin by extracting the $L$-hop neighbors of the query nodes in $V_T$, since \gnns typically rely on at most $L$-hop neighborhood information for each node. Additionally, we globally share the extracted subgraph, the \gnn model, and the set of explainability measures (line~\ref{pseudo-pre}).
Next, we perform the workload partitioning using procedure \cp (line~\ref{pseudo-cp}), which clusters the query nodes and assigns each partition to a separate thread. Within each thread, we first perform influential node prioritization (lines~\ref{pseudo-order1}–\ref{pseudo-order2}). Then, following the resulting prioritization order, we process each node and update the shared edge information table to reuse previously computed results (lines~\ref{pseudo-eist1}–\ref{pseudo-eist2}).
Finally, we aggregate the skyline explanations computed by all threads and return the consolidated results for the entire query batch (lines~\ref{pseudo-final1}–\ref{pseudo-final2}).

\begin{example}
As illustrated in Figure~\ref{fig:para-frame}, 
six \sgxq queries have been issued with 
specified output nodes $v_1$ to $v_6$. 
Following the intra-similarity analysis (procedure \cp ), the query workload is partitioned into three 
clusters $\{v_1, v_3\}$, $\{v_4, v_5\}$ 
and $\{v_2, v_6\}$ (the latter two are not shown), by maximizing the intra-cluster Jaccard similarity of their $2$-hop neighbors. The nodes in the same cluster are sorted by their core numbers by adapting $k$-core decomposition algorithm~\cite{core_decomp}. 
In this case, $v_1$ is scheduled to be processed first, followed by $v_3$. 
During the processing, all threads consistently invoke \apxsx, yet 
share the edge information asynchronously, 
via the shared edge information table. For 
example, 
once the edges in the neighbors of $v_1$ are verified, its auxiliary information  
is shared 
for fast local computation without 
recomputation. For instance, 
$e_{17}$ is skipped for $v_3$ since it has 
$e_{17}$ in its neighbor graphs. 
\end{example}

\stitle{Parallel Time Cost}. 
The time cost of \parasx consists of two parts:
(1) The time cost of the cluster-based partitioning, which is:
$O\left(|V_T| \left(\max(G^L(v)) + \log m\right)\right)$,
where $|V_T|$ is the number of test nodes, $G^L(v)$ denotes the $L$-hop neighborhood of node $v$, and $m$ is the number of threads.
(2) The time complexity of the skyline explanation computation, which is:
$O\left(\frac{|V_T|}{m} |\zeta'| \left(\left(\log\frac{r_\Phi}{\epsilon}\right)^{|\Phi|} + L|G^L(V_T)|\right)\right)$,
where $|\zeta'|$ is the number of candidate explanations, $|\Phi|$ is the feature dimension, $r_\Phi$ is the feature range, and $|G^L(V_T)|$ is the average $L$-hop neighborhood size across all test nodes.
Combining the two, the total parallel time complexity for \parasx is:
\[
\resizebox{0.48\textwidth}{!}{
$O\left(|V_T| \left(\max(G^L(v)) + \log m\right) + \frac{|V_T|}{m} |\zeta'| \left(\left(\log\frac{r_\Phi}{\epsilon}\right)^{|\Phi|} + L|G^L(V_T)|\right)\right)$
}
\]
The main efficiency gain comes from the term $G^L(V_T)$: since each edge is computed only once and shared across relevant nodes, the overall complexity is proportional to the size of the $L$-hop neighborhood of the entire query set $V_T$. In contrast, without edge information sharing, the total cost would scale with the sum of the $L$-hop neighborhoods of individual nodes, i.e., $\sum_{v_t \in V_T} G^L(v_t)$.

\begin{figure}[tb!]
    \centering
\includegraphics[width=0.45\textwidth]{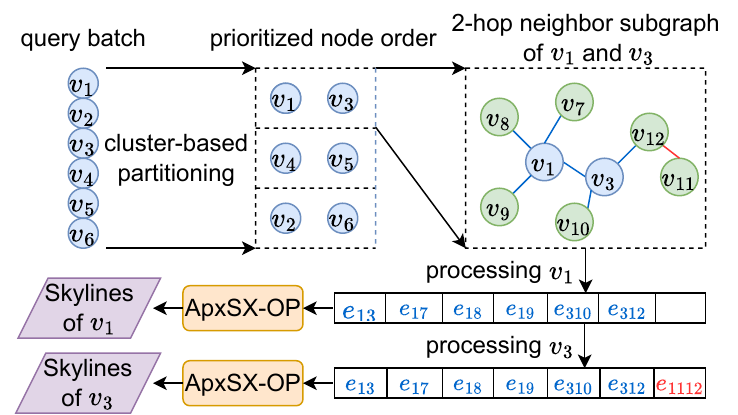}
    \caption{Running Example for \parasx. The blue edges are computed when processing $v_1$; The red edge is computed when processing $v_3$.
    When a query batch containing six nodes arrives, \parasx first partitions the queries into threads by maximizing intra-cluster Jaccard similarity. Within each thread, the queries are then ordered using the $k$-core algorithm.
    Consider a thread where the query order is $v_1$ followed by $v_3$. \apxsx initializes the shared edge information table by computing the edge scores (depicted in blue) from the neighborhood graph of $v_1$. During the processing of $v_3$, only one additional edge—$e_{1112}$ (shown in red)—requires computation, as the remaining edges are shared with $v_1$ and can directly reuse the values stored in the shared table.}   
     \vspace{-1ex}
    \label{fig:para-frame}
\end{figure}

\subsection{Additional Experimental Results}~\label{addition_exp_results}

\vspace{-1ex}
We present additional experimental study and case analysis. 

\stitle{Impact of Factors. }
We report  the impact of critical factors, i.e., size of explanations $k$, \gnn classes, and the number of layers $L$, on the effectiveness of 
generated skyline explanation over the  
\cora\ dataset. 

\eetitle{Impact of $k$}. Setting 
$\M$ as \gcn-based classifier with $3$ layers, 
we vary $k$ from 1 to 25. 
Since our competitors are not configurable w.r.t. $k$, we show the IPF and IGD scores of their solutions (that are independent of $k$), along with our \apxsx, \apxsxi, and \divsx (which are dependent on $k$) in Figures~\ref{fig:ipf_k} and~\ref{fig:igd_k}. 
As $k$ becomes larger, all achieve better scores 
and can consistently generate 
explanations with higher 
quality. This is because larger $k$ 
allows more dominating subgraphs 
to be verified, whenever possible 
for at least one measure. 
In contrast, our competitors are not aware 
of the growth of $k$, given that 
they may ``stuck'' at local optimal 
explanations that are optimized over a single measure, hence 
fail to improve the solution 
even with larger $k$. 
Moreover, \moe\ does not explicitly constrain the size of the explanation set, which can result up to 200 explanatory subgraphs based on our empirical results.
Meanwhile, even with $k$=$1$, our methods outperform other competitors for both IPF and IGD. 

\eetitle{Varying \gnn classes}. Fixing $k$=$10$, 
we report IPF scores of $3$-layer \gnn explainers 
for representative \gnn classes including \gcn, \gat, and \gin. 
As verified in Figure~\ref{fig:ipf_gnn}, \apxsx, \apxsxi, and \divsx consistently 
outperform other explanations for different 
 \gnn classes. 
This verifies that our approaches 
remain stable in generating high-quality 
explanations for different types of \gnns. 
Our observations for IGD scores are consistent, 
hence omitted. 

\eetitle{Varying number of layers}. Fixing $k$=$10$, 
we report IGD scores of \gcn explainers, 
with the number of layer $L$ varies from 1 to 3. 
over \cora. 
Fig~\ref{fig:ipf_L} shows that \apxsx, \apxsxi, and \divsx 
are able to achieve better IGD scores for generating 
explanations with more complex \gnns having ``deeper'' 
architecture. Indeed, as $L$ becomes larger, 
our all three methods are able to identify more
connected subgraphs that contribute to 
better $\fplus$ and $\fminus$ scores, hence 
improving explanation quality.

\begin{figure}[tb!]
    \centering
    \includegraphics[width=0.45\textwidth]{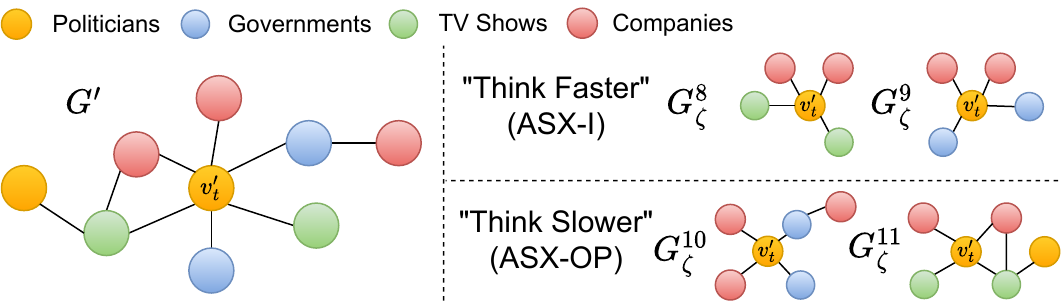}
    \caption{Different edge explore strategies show different preferences for returned skyline explanation.}
    \label{fig:fastslow}
\end{figure}

\stitle{Case Study: Edge Growing v.s. Onion-Peeling.}
In this case study, we compare the application scenarios 
of \apxsx and \apxsxi, as shown in Figure~\ref{fig:fastslow}. We observe that \apxsx, 
with onion peeling, cater users' needs who 
``think slow'' and seeks in-depth  
explanations that are counterfactual.
Such larger and counterfactual explanations are more likely to affect \gnn behaviors 
if removed~\cite{liu2021multi}.    
\apxsxi, on the other hand, due to its 
containment of edges 
closer to targeted ones via edge insertions/growths, tends to produce  --
at its early stage -- 
smaller, factual explanations as convincing evidences. 
This caters users who 
``think fast'' and is good at  
quick generation of factual explanations~\cite{liu2021multi}.
For example, in answering ``why'' 
node $v'_t$ is recognized as a ``Politician'' site, 
$G_{\zeta}^{8}$ and $G_{\zeta}^{9}$ are among skyline 
explanations from \apxsxi at early stage, 
revealing the fact that $v'_t$
frequently interact with TV shows, Governments, and 
Companies, covering its ego-networks 
as quick facts. $G_{\zeta}^{10}$ and $G_{\zeta}^{11}$, 
which are reported by \apxsx in its early 
stage via “onion-peeling”, are relatively larger counterfactual 
explanations that suggest  
interaction patterns with more 
complex structures. 
For example, $G_{\zeta}^{11}$ includes two-hop neighbors and a triangle, which indicates the importance of the corresponding company and TV show. Most importantly, another politician's two-hop neighbor also shares these nodes, revealing the close relationship between the two politicians.

\end{document}